\documentclass[lettersize,journal]{IEEEtran}
\usepackage{amsmath,amsfonts}
\usepackage{algorithmic}
\usepackage{algorithm}
\usepackage{array}
\usepackage[caption=false,font=normalsize,labelfont=sf,textfont=sf]{subfig}
\usepackage{textcomp}
\usepackage{stfloats}
\usepackage{url}
\usepackage{verbatim}
\usepackage{graphicx}
\usepackage{cite}
\usepackage{float}
\usepackage{booktabs} 
\usepackage{longtable}
\usepackage{stfloats}
\usepackage{xspace}
\usepackage{tikz}
\usepackage{flushend}
\usepackage{hyperref}
\usepackage{multirow} 
\usepackage{graphicx} % 引入图形处理的包
\usepackage{subcaption} % 引入子图的包
\usepackage{tabulary,xspace}

\begin{document}

\title{Low-Rank Adaption on Transformer-based Oriented Object Detector for Satellite Onboard Processing of Remote Sensing Images}

%\author{IEEE Publication Technology,~\IEEEmembership{Staff,~IEEE,}
        % <-this % stops a space

\author{Xinyang Pu,~\IEEEmembership{Graduate Student Member,~IEEE},~Feng Xu,~\IEEEmembership{Senior Member,~IEEE}

\thanks{This work was supported by the Science and Technology Commission of Shanghai Municipality (23JC1400501). (Corresponding author: Feng Xu). The authors are with the Key Laboratory for Information Science of Electromagnetic Waves (Ministry of Education), School of Information Science and Technology, Fudan University, Shanghai 200433, China. (fengxu@fudan.edu.cn)}% <-this % stops a space

\thanks{Manuscript received June 4, 2024; revised ***.}}
\markboth{Journal of \LaTeX\ Class Files,~Vol.~14, No.~8, April~2024}%
{Shell \MakeLowercase{\textit{et al.}}: A Sample Article Using IEEEtran.cls for IEEE Journals}

\IEEEpubid{0000--0000/00\$00.00~\copyright~2024 IEEE}

\maketitle

\begin{abstract}
Deep learning models in satellite onboard enable real-time interpretation of remote sensing images, reducing the need for data transmission to the ground and conserving communication resources. As satellite numbers and observation frequencies increase, the demand for satellite onboard real-time image interpretation grows, highlighting the expanding importance and development of this technology. However, updating the extensive parameters of models deployed on the satellites for spaceborne object detection model is challenging due to the limitations of uplink bandwidth in wireless satellite communications. To address this issue, this paper proposes a method based on parameter-efficient fine-tuning technology with low-rank adaptation (LoRA) module. It involves training low-rank matrix parameters and integrating them with the original model's weight matrix through multiplication and summation, thereby fine-tuning the model parameters to adapt to new data distributions with minimal weight updates. The proposed method combines parameter-efficient fine-tuning with full fine-tuning in the parameter update strategy of the oriented object detection algorithm architecture. This strategy enables model performance improvements close to full fine-tuning effects with minimal parameter updates. In addition, low rank approximation is conducted to pick an optimal rank value for LoRA matrices. Extensive experiments conducted on the DOTAv1.0, HRSC2016, and DIOR-R datasets verify the effectiveness of the proposed method. By fine-tuning and updating only 12.4$\%$ of the model's total parameters, it is able to achieve 97$\%$ to 100$\%$ of the performance of full fine-tuning models. Additionally, the reduced number of trainable parameters accelerates model training iterations and enhances the generalization and robustness of the oriented object detection model. The source code is available at: \url{https://github.com/fudanxu/LoRA-Det}.
\end{abstract}

\begin{IEEEkeywords}
Oriented object detection, parameter efficient fine-tuning (PEFT), low rank adaption (LoRA)
\end{IEEEkeywords}

\section{Introduction}
% \IEEEPARstart{T}{he} Parameter-Efficient Fine-Tuning (PEFT) technology \cite{PEFT} enables the training and updating of only a small fraction of the model's parameters, thereby enhancing the model's performance, and in some cases, even closely approximating the effects of full fine-tuning, and on the other hand, making fine-tuning large model possible under resource constraints. The PEFT techniques have been widely explored for applications of foundation models for their typical advantages on reducing training cost and hardware resource requirements, such as Large Language Models (LLMs) \cite{Llama}, while the application of PEFT to vision foundation models for flexible downstream tasks is still in the early stages of development, with only a few tasks, including segmentation with the Segment Anything Model (SAM) \cite{sam} and generation applications with diffusion models \cite{DDPM}, employing PEFT. 

\IEEEPARstart{A}{s} artificial intelligence technology advances, object detection models based on deep learning are playing an increasingly important role in the task of remote sensing image interpretation. The current approach typically involves transmitting images captured by satellites to ground stations, where deep learning models then perform inference and interpretation. However, processing massive volumes of remote sensing images through ground station models places a significant burden on satellite communication transmission and results in noticeable interpretation delays. This issue is exacerbated by the rapid increase in the number of satellites and the enhanced frequency of remote sensing observations, supported by advancements in aerospace technology. Therefore, satellite onboard models provide a reasonable potential for effective remote sensing image interpretation by directly processing the real-time captured images on each satellites without image transmission.

Moving forward, it is anticipated that onboard satellite deep learning models will be frequently updated to enhance both performance and generalization after retraining with continuous data streams. However, the full updating of all model parameters presents a considerable challenge and is time-intensive due to the limited uplink bandwidth available in satellite communications, as depicted in Figure 1. Consequently, we aim to develop a sophisticated algorithm that necessitates updating only a subset of the model's parameters to achieve enhanced detection performance. After retraining the detection model with an abundance of new remote sensing image data on a ground platform, only this subset of parameters will be modified and scheduled for upload to the onboard satellite model via uplink communication. This method is specifically designed to boost model performance while simultaneously minimizing the bandwidth requirements for uplink transmissions in satellite communications.

\IEEEpubidadjcol 

\begin{figure}[!ht]
\centering
\includegraphics[width=3.5in]{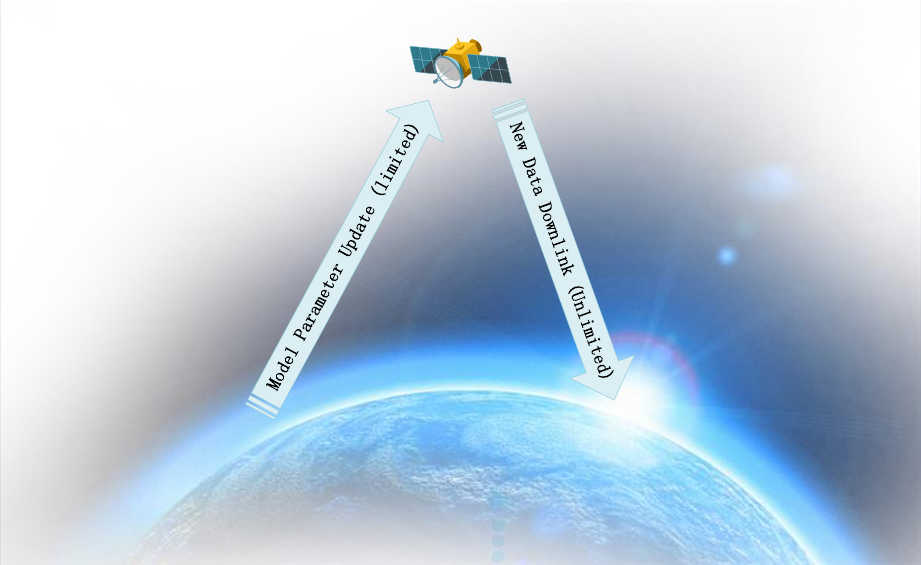}
\caption{The transmission limitations in the uplink bandwidth of spaceborne object detection models.}
\label{fig1}
\end{figure}

We are inspired by the Parameter-Efficient Fine-Tuning (PEFT) concept for large models, which allows for the training and updating of only a small fraction of the model's parameters. PEFT technology enhances the model's performance and can approximate the effects of full fine-tuning, while also enabling fine-tuning of large models under resource constraints. The PEFT techniques are widely used in foundation models such as Large Language Models (LLMs) \cite{Llama} to reduce training costs and hardware requirements. The application of PEFT to vision foundation models for flexible downstream tasks is still in its early stages. A few tasks, including segmentation applications with the Segment Anything Model (SAM) \cite{sam} and generation applications with diffusion models \cite{DDPM}, have already employed PEFT.

% Inspired by the PEFT concept for large models, our goal, adopting the low rank adaption (LoRA) module \cite{LoRA}, aims to fine-tune the spaceborne object detection model with continuously incoming new remote sensing data, using as minimal a volume of trainable parameters as possible, without altering the model structure. Furthermore, in the context of uplink bandwidth limitations for satellite communications as shown in Figure 1, the portion of parameters that require updating is to be transmitted to the weights of object detection model onboard the satellite payload and integrated. This is the reasoning for utilizing the PEFT techniques for training the oriented object detection models in remote sensing images, aiming to enhance the detection capabilities and generalizability.
To address the challenge of satellite onboard model parameter updating, we intend to adopt the low rank adaption (LoRA) module \cite{LoRA} to fine-tune the satellite onboard object detection model as a PEFT strategy. We use continuously incoming new remote sensing data, employing the minimal volume of trainable parameters possible without altering the model structure. Furthermore, only these trainable parameters that require updating are transmitted and then integrated into the weights of the object detection model onboard the satellite payload. This explains the use of PEFT techniques for training oriented object detection models in remote sensing images. The aim is to enhance the detection capabilities and generalizability of these models.

% However, investigation and development of PEFT technology in object detection algorithms remain insufficient \cite{AiRs}, attributed to the non-existence of a foundational model in object detection and the complexity of detection algorithm architectures. On the other hand, while most PEFT methods are applied to the backbone of deep learning networks and have been extensively studied, there is a lack of investigation concerning the full scope of the algorithmic architecture for diverse downstream tasks. 

The existing investigation and development of PEFT technology in object detection algorithms is not sufficient \cite{AiRs}. This is due to the lack of a foundational model in object detection and the complexity of detection algorithm architectures. Most PEFT methods that have been extensively studied are applied to the backbone of deep learning networks. However, there is a lack of investigation concerning the full scope of the algorithm architecture for diverse downstream tasks.

% Common LLMs rely on transformer frameworks \cite{Transformer}, where backbone weights constitute the majority of the parameters, leading to prevalent PEFT algorithms being primarily explored within the attention and MLP (Multi-Layer Perceptron) modules of a Transformer-based backbone. Similarly, the LoRA method is employed on the transformer module for fine-tuning stable diffusion models towards a variety of image generation applications by manipulating representations in the latent space. Nevertheless, research on parameter-efficient fine-tuning techniques for object detection models is still limited, particularly in terms of exploring fine-tuning beyond the backbone structures.
Furthermore, PEFT module designs of network structure mostly focus on common LLMs with transformer frameworks \cite{Transformer}, where backbone weights constitute the majority of the parameters. This has led to prevalent PEFT algorithms being primarily explored within the attention and Multi-Layer Perceptron (MLP) modules of a Transformer-based backbone. Similarly, the LoRA method is employed on the transformer module for fine-tuning stable diffusion models \cite{DDPM}. This approach manipulates representations in the latent space to accommodate a variety of image generation applications. Nevertheless, there is still limited research on parameter-efficient fine-tuning techniques for object detection models based on convolutional neural networks. Specifically, there is a gap in studies focusing on fine-tuning parts beyond the backbone structures.

In summary, we present three challenges of PEFT application on object detection task:
\begin{enumerate}
\item {Object detectors have more complicated architectures and specially-designed modules, compared with image-level classification and pixel-level segmentation. Specifically, the hierarchical feature map and region proposal generation add complexity to the design of PEFT methods. Additionally, the fine-grained detection head blocks present further challenges.}
% \item {Focusing on our spaceborne scenarios, satellite hardware imposes limitations on the parameter quantities and computing resources of the detectors. Consequently, we delve into the application of PEFT on normal-sized models (with millions of parameters) \cite{orcnn}, for which some empirical principles of PEFT on large models are no longer feasible.}

\item {For spaceborne onboard processing scenarios, satellite hardware imposes limitations on the parameter quantities and computing resources of the detectors. Consequently, we are interested in the application of PEFT on normal-sized models instead of large models, which contain few millions of parameters \cite{orcnn}. It is found that some empirical principles of PEFT on large models are no longer feasible for these models.}

\item {Selecting a proper low-rank value for trainable matrices in LoRA module on normal-sized models is both critical and challenging. This is because the intrinsic rank may be relatively high, which differs from the hypothesis that matrices in large models tend to contain low intrinsic dimensionality \cite{SAID}.}
\end{enumerate}

To tackle these challenges, this paper explores the architectural design of parameter-efficient fine-tuning techniques in the context of oriented object detection algorithms. It involves with identifying the layers that are most important for fine-tuning and determining the optimal structures for hybrid PEFT methods to leverage new data in enhancing the model's detection capabilities. In addition, it includes figuring out a reasonable low-rank value for each matrix in LoRA modules to optimize the trade-off between model performance and resource usage. 
% It involves identifying the layers most impactful for fine-tuning, determining the optimal structures for hybrid PEFT methods to leverage augmented data in enhancing the model's detection capabilities and figuring out reasonable low rank value for each matrix on LoRA modules for optimal trade-off between model performance and resources.

The main contributions of this paper can be summarized as follows: 
\begin{enumerate}
\item {The proposed method designs a novel parameter efficient fine-tuning algorithm for oriented object detection named LoRA-Det, conducting Low Rank Adaption module through the whole framework including transformer-based backbone and detection head. }
\item {The necessity and effective strategies for adopting PEFT in comprehensive modules of the detection network are explored. A hybrid PEFT and full fine-tuning strategy in transformer and convolutional layers is proposed to achieve a satisfactory trade-off between detection performance and the quantity of trainable parameters.}
\item {A low rank approximation method is employed to find a proper rank $r$ for the trainable matrices of LoRA modules. Extensive experiments are conducted on multiple remote sensing image datasets to verify the effectiveness and rationality of the proposed algorithm.}
\end{enumerate}

The rest of this paper is arranged as: Section II presents an extensive review of prior studies concerning LoRA-Det. Section III elaborates on the detailed architecture of the proposed method LoRA-Det. Section IV presents a thorough analysis of quantitative and qualitative evaluation results, comparing multiple experiments and conducting an ablation study. Finally, Section V concludes the paper.

\section{Related works}
\subsection{Parameter Efficient Fine-Tuning}
Parameter efficient fine-tuning technologies are widely utilized in large model training and offer pragmatic solutions for efficiently adapting large models to various downstream tasks. Essentially, PEFT involves fine-tuning the parameters of a pre-trained large model to suit a specific task or domain, while minimizing the introduction of additional trainable parameters or computational resources. The emergence of PEFT technology originated from large language models in natural language processing, and subsequently began to be gradually applied in the field of computer vision, including Vision Transformer and diffusion models. 

Serial Adapter \cite{SerialAdapter} introduces additional trainable parameters by inserting two light-weight adapter layers into the transformer blocks of original networks, one after the self-attention layer and the other after the FFN layer. AdapterFusion \cite{Adapterfusion} improves the computation efficiency by adding adapter layers after 'Add \& Norm'. Parallel adapter (PA) \cite{PA} proposes a parallel side-network structure of adapters on transformer blocks, similarly as CIAT \cite{CIAT} and KronA \cite{KronA}. Such adapter-based methods add extra layers to original model architectures and some of them conduct multi-task learning such as AdaMix \cite{AdaMix} and AdapterSoup \cite{AdapterSoup}.

Another way of PEFT based on low rank adaption (LoRA) introduces no inference latency for model by updating existing parameters through matrix multiplication. LoRA is the most widely recognized reparameterization technique of PEFT. DyLoRA \cite{DyLoRA} proposes a strategy to select proper rank for LoRA module and reduce the time cost of training process. AdaLoRA \cite{AdaLoRA} leverages singular value decomposition (SVD) to obtain three weight matrices of LoRA module for fine-tuning. Multiple research are focused on improving performance of LoRA via diverse approaches including Laplace approximation \cite{Bayesian} and novel training strategies \cite{LoRA+} \cite{LoraHub}. DoRA \cite{DoRA} designs Weight-Decomposed Low-Rank Adaptation by decomposing model weights into magnitude and direction and introduces a novel fine-tuning strategy for weight matrices. In addition, various hybrid combination of PEFT strategies are explored to enhance the fine-tuning effectiveness such as UniPELT \cite{UniPELT} and LLM-Adapters \cite{LLM-Adapters}. 

\subsection{Low Rank Adaption in Computer Vision}
The application of PEFT technology in various vision tasks and vision-language models \cite{T2I-Adapter} \cite{Tip-Adapter} has also been widely investigated. Considering LoRA focus on fine-tuning the $Q$,$K$,$V$ matrices of parameter weight in attention modules, this approach is mainly explored as a fine-tuning strategy in the Vision Transformer (ViT) \cite{ViT} backbone of large models, such as Segment Anything Model (SAM) and diffusion models. LoRAT \cite{LoRAT} leverages LoRA module to fine-tune transformer encoder of ViT-g backbone and designs a multilayer perceptron (MLP)-based head to improve model performance and reduce training cost for visual object tracking. \cite{MED-PEFT} delves into LoRA method for PEFT technologies on medical vision foundation models. In \cite{mipr}, LoRA is utilized on pre-trained ViT backbone and Single Center Loss (SCL) is conducted to enhance generalization capability of face forgery detection model. The introduction of Concept Sliders \cite{ConceptSliders} presents plug-and-play LoRA adapters, facilitating precise adjustments to concepts (e.g., age, smiling) within a diffusion model. A novel semantic segmentation model is proposed by \cite{10412208}, fine-tuning ViT-based image encoder of Segment Anything Model (SAM) with low-rank adaptation approach for feature extraction on optical aerial images. AiRs \cite{AiRs} explores diverse PEFT technologies, such as LoRA, adapter and bitfit, on foundation models and designs adapter-based modules for multiple remote sensing downstream tasks. 

\subsection{Oriented Object detection}
Oriented object detection, a popular topic in computer vision, has garnered comprehensive attention in both algorithmic advancements and practical applications. In response to the challenges posed by densely distributed objects with arbitrary orientations in remote sensing imagery, numerous algorithms have been investigated to accurately locate rotated bounding boxes encompassing object edges on publicly available aerial image datasets from satellites, such as DOTA \cite{DOTA}, HRSC2016 \cite{HRSC}, and DIOR-R \cite{DIOR}, among others. The RoI Transformer \cite{RoITransformer} incorporates the RRoI learner and Rotated Position-Sensitive RoI Align. The former learns the transformation from Horizontal RoIs to Rotated RoIs, while the latter extracts rotation-invariant features from RRoI to facilitate oriented object detection tasks. Subsequently, the Rotation-equivariant Detector (ReDet) \cite{ReDet} integrates rotation-equivariant networks into the backbone and utilizes Rotation-invariant RoI Align (RiRoI Align) to extract rotation-invariant features, leading to significant improvements in detection performance for aerial images with oriented bounding boxes. The gliding vertex \cite{Gliding} introduces an effective representation of oriented bounding boxes to mitigate detection errors and confusion issues. The Oriented R-CNN \cite{orcnn} utilizes an oriented Region Proposal Network (oriented RPN) and Oriented R-CNN head to achieve state-of-the-art detection accuracy on two public datasets, enabling the acquisition of high-quality oriented proposals and refined oriented Regions of Interest. The R3Det \cite{R3Det}, an efficient and fast single-stage detector, effectively resolves the feature misalignment problem for large aspect ratio object detection and introduces the SkewIoU loss to accurately estimate object orientations. RTMDet \cite{rtmdet} proposes an efficient light-weighted real-time object detector and is utilized on oriented object detection in aerial images.

\section{METHODOLOGY}
\subsection{Low Rank Adaption in LoRA-Det} 

\subsubsection{\textbf{Preliminaries of LoRA}}
Deep neural networks contain massive parameters for dense matrix multiplication. The parameter matrices in a well pre-trained weight tend to have full-rank, while updating parameters by training new data under similar scene for an identical task, the updated matrices can be restricted in low dimension to decrease the number of updated parameters by low-rank adaption (LoRA) \cite{LoRA}. 

\begin{figure}[!ht]
\centering
\includegraphics[width=3.2in]{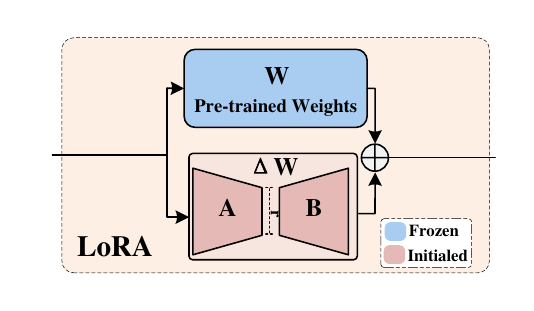}
\caption{The structure of LoRA module.}
\label{fig4}
\end{figure}

As shown in Figure 2, to update a parameter matrix from $ W \in \mathbb{R}^{d \times k} $ in the pre-trained weight where $d \times k$ denotes the dimensions of the matrix $W$ to $W’ \in \mathbb{R}^{d \times k}$ as:
\begin{equation}
    {W’=W + \Delta W} 
    \label{eqn:attention}
\end{equation}
where $\Delta W \in \mathbb{R}^{d \times k}$, LoRA decomposes $\Delta W$ to two low dimension matrices A and B as:

\begin{equation}
    {\Delta W = BA }
\end{equation}
where $B \in \mathbb{R}^{d \times r}, A \in \mathbb{R}^{r \times k}$. By utilizing LoRA fine-tuning, the number of learning parameters is converted from $d \times k$ to $r \times (d+k)$. By adjusting the rank $r$, the number of learning parameters is reduced when $r < (d \times k) / (d+k)$, %$r < \frac{d \times k}{d+k}$
and the quantity of fine-tuned trainable parameter can be extremely reduced if $\text{rank } r \ll \min(d, k)$ . Therefore, the feature extraction of forward propagation in LoRA block can be described as:

\begin{align}
f = W'x = (W+\Delta W)x = (W+BA)x = Wx + BAx
\end{align}

During the training process, weight matrix W is frozen and does not require gradient calculation. Initialized matrix A follows random Gaussian distribution as $A = \mathcal{N}(0, \sigma^2)$ and matrix B is zero as $B = \mathbf{0}$, and these two matrices contains trainable parameters for gradient update and fine-tuning. Therefore, $\Delta W = BA$ is zero so that it will not affect the feature extraction at the beginning of model training. In addition, $W$ and $\Delta W$  have the same matrix dimension and can be summed coordinate-wise directly as:
\begin{equation}
W’= W+\Delta W =W+BA
\end{equation}
so LoRA maintains the original framework unchanged and brings no extra inference cost and latency after wight integration.

\begin{figure*} [!hbt]
\centering 
\includegraphics[width=\textwidth]{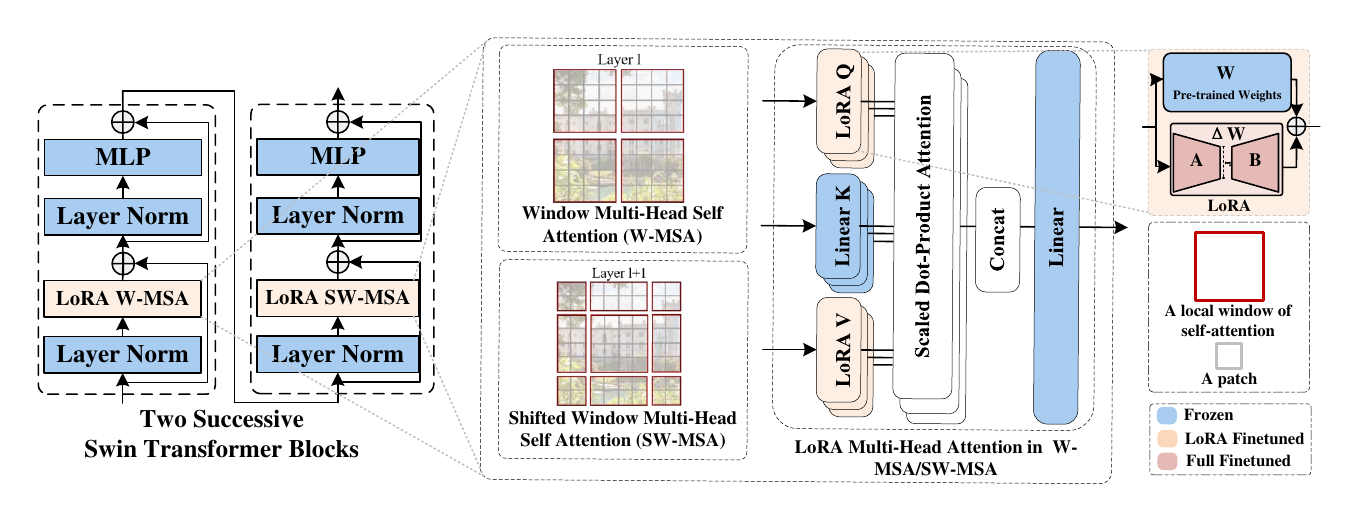}\\ 
\caption{The architecture of LoRA module in Swin Transformer block.}\label{fig_all} 
\end{figure*}

\subsubsection{\textbf{LoRA in Dense Blocks}}
LoRA can be employed in diverse structures of dense layers in neural networks by simply replacing the dense matrices in the pre-trained weights. The proposed method designes two types of LoRA-based PEFT modules for oriented object detection algorithm, including LoRA in linear layers and LoRA in Swin Transformer blocks.

For linear layers, the low rank matrices $A$ and $B$ are directly added in parallel next to the parameter matrix of the fully connected layer as illustrated in the Figure 2. 

For Swin Transformer blocks \cite{Swin}, the LoRA fine-tuning structure is depicted in Figure 3. In two successive Swin Transformer blocks, the image feature is partitioned evenly without overlapping into $8 \times 8$ patches and each $4 \times 4$ patches is regarded as a local window where the multi-head self attention is computed. As a result, $2 \times 2$ windows are built for multi-head self attention in a Swin Transformer block. For the first Swin Transformer block, the windows of size $4 \times 4$ patches are partitioned regularly from top-left to right-bottom, while in the second block, the start of window partitioning is shifted by $(2,2)$ patches and an efficient batch computation method named cyclic shifting is deployed to restore four full local windows and avoid additional attention computation from extra sub-windows.

After the window partition, the multi-head self attention with relative position bias is calculated in each local window as: 
\begin{equation}
    {\rm Attention}({Q}, {K}, {V}) = {\rm Softmax}(\frac{{Q}{K}^\mathsf{T}}{\sqrt{\rm d_{head}}}+B){V}.
    \label{eqn:attention}
\end{equation}
where $Q, K, V \in \mathbb{R}^{M^2\times d}$ are the query, key and value matrices; $d_{head}$ is the query/key dimension, $B \in \mathbb{R}^{M^2 \times M^2}$ and $M^2$ is the number of patches in a window.

We utilize LoRA layer to adapt the parameter matrices $W_q$ and $W_v$ in the attention module of Swin Transformer blocks following the empirical experiments in LoRA\cite{LoRA}, and other layers in the Transformer blocks are frozen during the fine-tuning process.

Therefore, by employing this approach, which leverages the LoRA module for fine-tuning the backbone, there will be a substantial reduction in the quantity of trainable parameters. Depending on the value of rank $r$, it is reasonable to adjust the amount of parameters necessitated for training. This enables the model to be flexibly adapted to new data domains.

\subsection{Parameter Efficient Fine-tuning in Oriented Object Detection}

\begin{figure*} [!bht]
\centering 
\includegraphics[width=\textwidth]{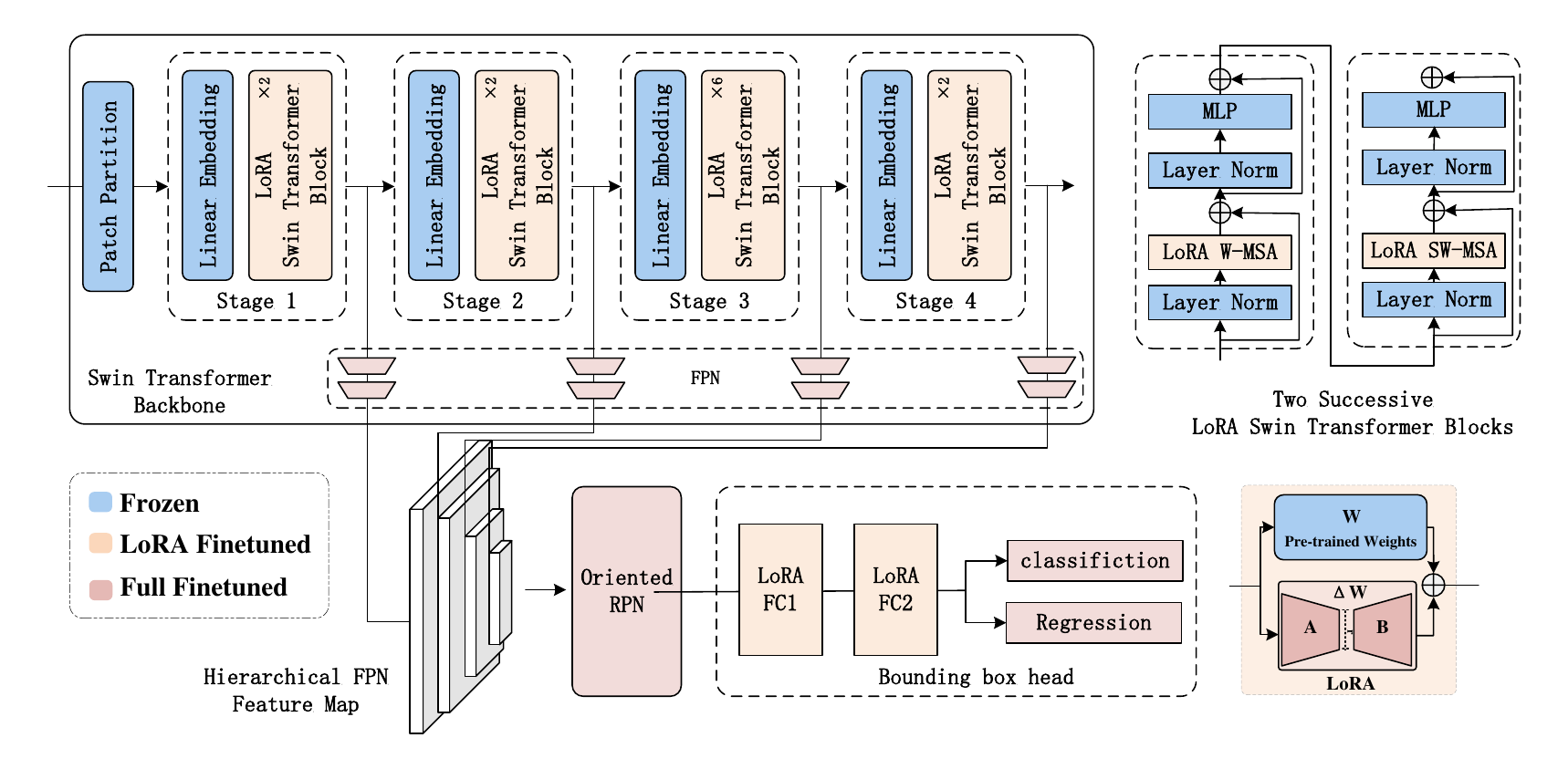}\\ 
\caption{The architecture of the proposed LoRA-Det.}\label{fig_all} 
\end{figure*}

As an oriented object detection network shown in Figure 4, the Oriented RCNN \cite{orcnn} is fundamentally comprised of the backbone based on Swin-Transformer, Feature Pyramid Network (FPN), oriented Region Proposal Network (RPN) head, and bounding box head. An analysis of the parameter quantity distribution, as depicted in Figure 5, reveals that the backbone and detection head (named as bounding box head in Oriented RCNN) constitute the bulk of the parameters, serving the critical functions of feature extraction and detection regression and classification, respectively. Therefore, the LoRA technique is employed within both the backbone and bounding box head components as shown in Figure 4. This strategy not only diminishes the volume of parameters necessitating training and expedites the training process but also leverages parameter-efficient fine-tuning methodologies to substantially improve the model's detection capabilities across new data domains.

\begin{figure}[!ht]
\centering
\includegraphics[width=3.5in]{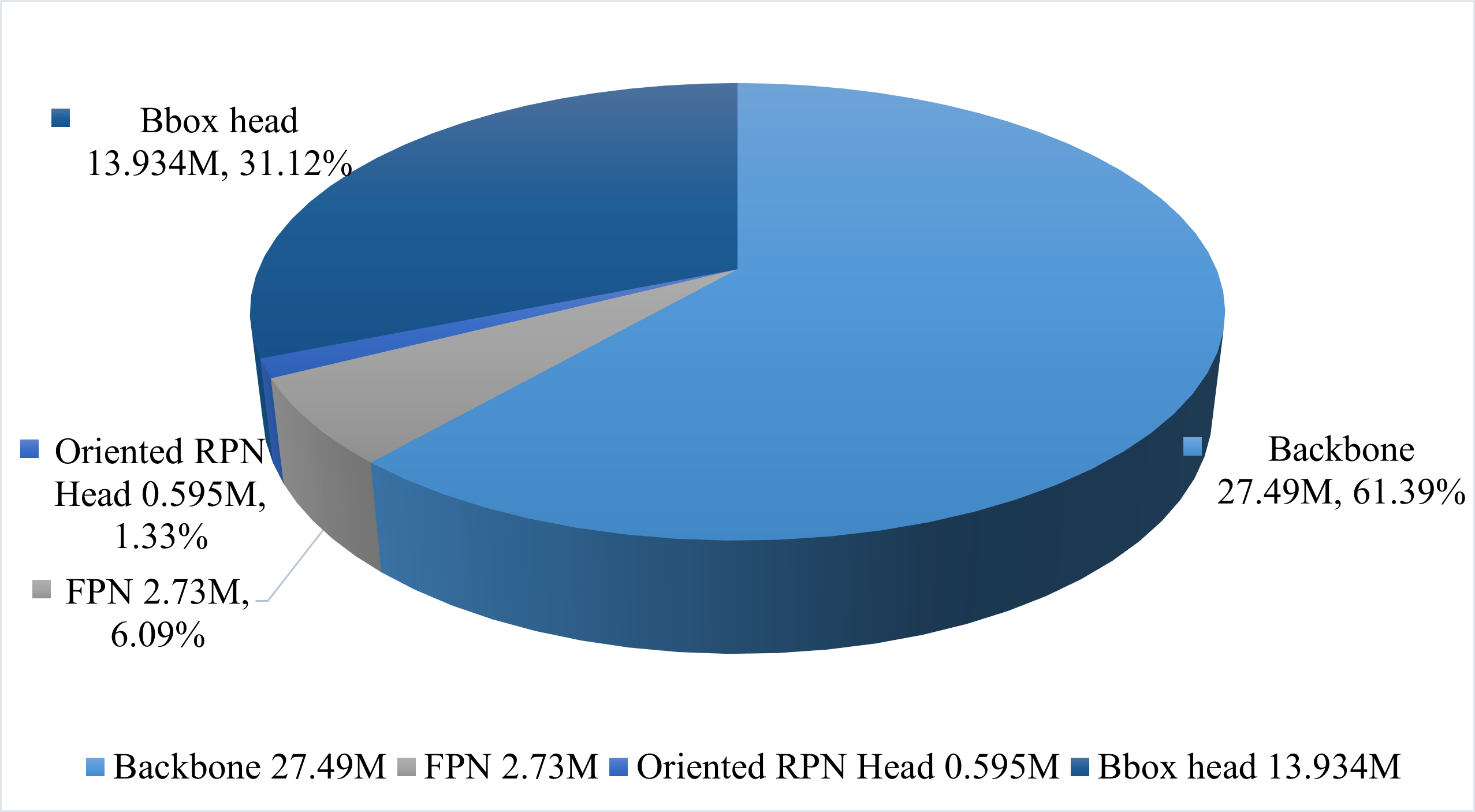}
\caption{Parameter distribution of oriented object detection model Oriented RCNN.}
\label{fig4}
\end{figure}

\subsubsection{LoRA Fine-tuning on Transformer-based Backbone}
In the Swin Transformer backbone, LoRA is applied within the attention modules of the transformer blocks. As illustrated in Figure 4, an input image, after undergoing a non-overlapping patch partition, sequentially passes through four stages to extract image features. Stage 1 consists of a linear embedding and two Swin Transformer blocks, while stages 2, 3, and 4 each include a patch merging and multiple successive Swin Transformer blocks. Specifically, stages 2, 3, and 4 are equipped with two, six, and two Swin Transformer blocks, respectively. The structure of every two consecutive Swin Transformer blocks, as shown in Figure 3, integrates the LoRA module within the query and value matrices of the Window Multi-head Self-Attention (W-MSA) and Shifted Window Multi-head Self-Attention (SW-MSA) modules following the strategies described in the previous section.

\subsubsection{Hybrid Fine-tuning in Oriented RCNN}
In the Oriented RCNN framework, the efficient fine-tuning of parameters in the detection head is realized through the adaptation of low-rank LoRA to the fully connected layers. After the Oriented RCNN head performs oriented Region of Interest (RoI) alignment on the oriented proposals, the resulting fixed-size feature vector is fed into the fully connected layers. LoRA module is applied to these two parameter-dense fully connected layers, utilizing low-rank matrices to fine-tune the shared fully connected layers of the detection head. 

The proposed algorithm employs parameter-efficient fine-tuning for the backbone and Oriented RCNN head using the LoRA module, as outlined earlier. The remaining components, primarily convolutional layers with fewer parameters, are subjected to standard full fine-tuning. Full fine-tuning approach encompasses modules such as the Feature Pyramid Network (FPN) and the Oriented Region Proposal Network (RPN) head. 

Moreover, to augment the detection and classification performance of the model, the lightweight classification and regression fully connected layers located at the terminal part of the Oriented RCNN detection head also employ the full fine-tuning strategy. This method ensures a balanced optimization across the network, enhancing the model's overall performance by fully adjusting both its core and auxiliary components.

\subsection{Low Rank Approximation}
To select an efficient rank $r$ value for each LoRA module in normal size detector to achieve reliable trade-off between quantities of trainable parameters and fine-tuning performance, low rank approximation method is explored on weight matrices of LoRA-Det. Inspired by the well-known Singular value decomposition (SVD) method on data compression, we analyse the intrinsic rank of parameter matrices and evaluate their low rank approximation errors to adopt optimal rank $r$ value for LoRA module as illustrated in Figure 6.

\begin{figure}[!ht]
\centering
\includegraphics[width=3.5in]{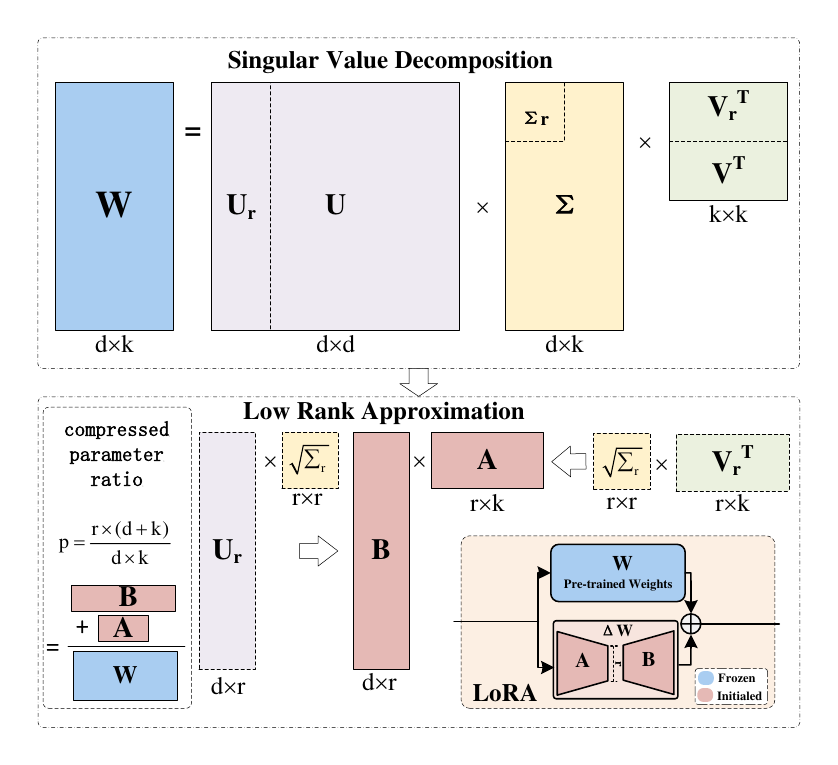}
\caption{Low rank approximation method in LoRA-Det.}
\label{fig4}
\end{figure}

For a weight matrix $W \in \mathbb{R}^{d \times k}$ in deep learning network and its rank $r\leq \rm{min}(d,k)$, if the matrix is from large model with huge dimensions, its intrinsic rank might be low as depicted in \cite{SAID}, whereas the normal size matrix's intrinsic rank tends to be higher, possibly even full rank as $r = \rm{min}(d,k)$. Singular value decomposition is applied on the weight matrix $W$ to obtain three matrices $U\in \mathbb{R}^{d \times d}$, $\Sigma\in\mathbb{R}^{d \times k}$ and $V\in \mathbb{R}^{k \times k}$ for low rank approximation as:

\begin{equation}
    {\rm SVD}(W) = U\Sigma V^T
    \label{eqn:SVD}
\end{equation}
where $U^TU=\mathbb{I}^{d \times d}$, $VV^T=\mathbb{I}^{k \times k}$ and $\Sigma$ is a real diagonal matrix with no negative elements. In matrix $\Sigma$, the singular values are sequenced in descending order. To conduct low rank $r$ in LoRA matrix, top-$r$ singular values are reserved for presenting most important parameter information and matrices $U$,$V$ are formalized to corresponding dimension as:
\begin{equation}
    \begin{cases}
    {U_r} = {U}_{[:, : r]}, \\
    {\Sigma_r} =  {\Sigma}_{[:r,:r]}, \\
    {V^T_r} =  {V^T}_{[:r, :]}, \\
    \end{cases}
\end{equation}
where these three truncated low-rank matrices act as $U_r\in \mathbb{R}^{d \times r}$, $\Sigma_r\in\mathbb{R}^{r \times r}$ and $V^T_r\in \mathbb{R}^{r \times k}$.

The low rank approximation approach aims to approximate the parameter matrix $W$  by a the rank-$r$ matrix $\bar W$ to minimize the euclidean distance between them. As matrix $W \in \mathbb{R}^{d \times k}$ and $r\leq \rm{min}(d,k)$, the constraints are formulated as below:
\begin{equation}
\begin{cases}
    \parallel \bar{W} - W \parallel_\infty \leq \epsilon, \\
     {\bar W} = U_r\Sigma_r V_r^T
\end{cases}    
\end{equation}
where $\epsilon$ represents the low rank approximation error. It is proven by \cite{Eckart1936TheAO} that when $r' \leq r \leq \rm{rank}(W) \leq \rm{min}(d,k) $:

\begin{equation}
 ||W-\bar W||_\infty \leq ||W-W'||_\infty \label{eq:rank-k}
\end{equation}
which leads to a serious dilemma between compression effectiveness and approximation precision if the weight matrix $W$ is close to full-rank, while the compression can be very efficient if there is an low intrinsic rank in matrix $W$. 

Therefore, in order to analyze the relationship between the value of low rank $r$, the effectiveness of low-rank decomposition, and low rank approximation error, we introduce a compressed parameter ratio $p \in [0,1]$. For a pre-trained parameter matrix $W \in \mathbb{R}^{d \times k}$, its low rank approximation approach introduces three matrices: $U_r\in \mathbb{R}^{d \times r}$, $\Sigma_r\in\mathbb{R}^{r \times r}$ and $V^T_r\in \mathbb{R}^{r \times k}$. After the matrix multiplication operation, it is possible to establish a correspondence between the two low-rank matrices $A \in \mathbb{R}^{r \times k}$ and $B\in \mathbb{R}^{d \times r}$ within the LoRA module:
\begin{equation}
\begin{cases}
     {\bar B} = U_r\Sigma_r^{\frac{1}{2}} \\
     {\bar A} = \Sigma_r^{\frac{1}{2}} V_r^T
\end{cases}    
\end{equation}
Through this LoRA module, the number of trainable parameters is reduced from $d\times k$ to $d\times r+r\times k$ when $r \leq (d \times k)/(d+k)$. The compressed parameter ratio is defined as:
\begin{equation}
 p = \frac{r \times (d+k)}{d \times k}  \label{eq:cpr}
\end{equation}
 
Furthermore, we investigate the correlation of the compressed parameter ratio $p$ with the low rank approximation error, which influences the selection of the rank value $r$ on LoRA module. 

Specifically, if the low-rank approximation error consistently decreases as $p$ increases, it implies a higher intrinsic dimensionality of the matrix. Consequently, opting for a smaller $r$ may hurt the model's performance. Conversely, if the low-rank approximation error remains consistently low regardless of variations in $p$, it suggests the presence of a low intrinsic rank within the matrix. In such instances, employing a lower rank $r$ within the LoRA module can achieve optimal fine-tuning performance while concurrently economizing on trainable parameters.

\section{EXPERIMENT RESULTS AND ANALYSIS}
\subsection{Datasets} 
To assess the efficacy of the proposed method, three remote sensing image datasets for oriented object detection are utilized for validation.

(1) DOTAv1.0 \cite{DOTA}: The DOTA dataset stands as a comprehensive collection for oriented object detection, encompassing 2806 aerial images and 188282 labeled instances with oriented bounding box annotations across 15 different object categories. These categories include Bridge (BR), Harbor (HA), Ship (SH), Plane (PL), Helicopter (HC), Small vehicle (SV), Large vehicle (LV), Baseball diamond (BD), Ground track field (GTF), Tennis court (TC), Basketball court (BC), Soccer-ball field (SBF), Roundabout (RA), Swimming pool (SP), and Storage tank (ST). The images are in high-resolution, ranging from $800\times800$ to $4000\times4000$ pixels. For the purpose of training, both training and validation subsets are utilized, while the test dataset is evaluated through the online submission of testing results to the DOTA evaluation server.

(2) HRSC2016 \cite{HRSC}: The HRSC2016 dataset is utilized for the fine-grained ship detection within 31 categories annotated by oriented bounding boxes, comprising 1061 images whose sizes vary between $300\times300$ and $1500\times900$ pixels. The dataset is divided into a training set of 436 images and a validation set of 181 images, both of which are employed in the training process, with the remaining 444 images are for testing purposes. To assess detection accuracy on HRSC2016, the evaluation leverages the mean average precision (mAP) metric, aligning with the evaluation standards set by PASCAL VOC 2007.

(3) DIOR-R \cite{DIOR}: The DIOR-R dataset utilizes the images from the previously established DIOR dataset \cite{DIORISPRS}. Oriented bounding boxes are annotated for each instance to facilitate tasks involving oriented object detection. The dataset encompasses 23,463 images and 192,518 instances, spanning a broad spectrum of scenes and 20 common object categories on remote sensing: Airplane (APL), Airport (APO), Baseball Field (BF), Basketball Court (BC), Bridge (BR), Chimney (CH), Expressway Service Area (ESA), Expressway Toll Station (ETS), Dam (DAM), Golf Field (GF), Ground Track Field (GTF), Harbor (HA), Overpass (OP), Ship(SH), Stadium (STA), Storage Tank (STO), Tennis Court (TC), Train Station (TS), Vehicle (VE) and Windmill (WM).

\subsection{Data Configuration}
In order to investigate the performance of parameter-efficient fine-tuning algorithms on new data, the experimental design for the three datasets is as follows: The test set is conducted for validation, while the entire training and validation sets were both utilized for model training and are hereafter referred to as the training data. Here, 50$\%$ of the training data, which is solely sourced from the training set to ensure no data leakage, is regarded as pre-training data to determine the lower bound of the model's detection performance. Conversely, the performance of models subjected to full fine-tuning using the entire training and validation sets constitutes the upper bound. The proposed method employs the LoRA-based parameter-efficient fine-tuning on the pre-trained model using all training data and analyzes this against the performance upper bound of full fine-tuning.

\begin{table*}[t]
\centering
\caption{Low-rank value $r$ configurations and compressed parameter ratios of LoRA modules in LoRA-Det.}
\begin{tabular}{lcccccc}
\toprule
 LoRA Module & \multicolumn{4}{c}{Backbone} &  \multicolumn{2}{c}{Detection head}     \\
 & Stage 1& Stage 2&  Stage 3&  Stage 4& FC 1 & FC 2 \\
\midrule
Matrix dimension & (96,96) &(192,192) & (384,384) &(768,768) & (1024,12544) & (1024,1024)  \\
Low rank $r$ & 48& 48 & 48 & 48 & 64 & 16  \\
Compressed parameter ratio $p$ & 1 & 0.5 &0.25 &0.125 &0.068 &0.03125  \\
\bottomrule
\end{tabular}
\end{table*}

\subsection{Parameter Settings}
In our experiments, we utilize two NVIDIA RTX 3090, with a batch size of 4, employing the Adam optimizer. The learning rate is set to 0.0001, and weight decay is configured at 0.05. Our data augmentation strategy is limited on random flipping, and training is conducted using a single scale approach. For the DOTA dataset, images are cropped into slices with size of $1024\times1024$ pixels using a stride of $800\times800$, and the model is trained for 12 epochs. The HRSC2016 and DIOR-R datasets undergo image resizing to $1024\times1024$ and $800\times800$ pixels, respectively, before being fed into the network for training, with the HRSC2016 dataset undergoing 36 epochs of training and the DIOR-R dataset 12 epochs.

\subsection{Baselines of Partial and Full Fine-tuning}
Models trained using 50$\%$ of the training data are considered as a lower bound for the performance of pre-trained models. The upper bound of performance is obtained by training all model parameters using the whole training data. We hypothesize that LoRA-Det will outperform the lower bound in detection performance and slightly lag behind the upper bound achieved through full fine-tuning. However, it will significantly reduce the number of trainable parameters compared to full fine-tuning, showcasing its excellent parameter-efficient fine-tuning technique. This achievement represents a favorable trade-off between detection performance and parameter efficiency.

\subsection{Evaluation of Both Fine-tuning Performance and Parameter Efficiency in LoRA-Det}
The comprehensive experiments on three remote sensing datasets explore the efficiency of the proposed method, along with the comparison of various fine-tuning strategies, including: fine-tuning the backbone only; fine-tuning the detection head only; adopting LoRA modules in backbone and full fine-tuning the whole detection head (referenced as LoRA in the table); the proposed LoRA-Det without full fine-tuning the convolutional layers of FPN and Oriented RPN modules, LoRA-Det with hybrid fine-tuning the FPN and Oriented RPN modules. The low-rank value $r$ in LoRA module of the proposed LoRA-Det is adjusted for optimizing model performance.

The detailed low rank configurations of each LoRA module of the proposed LoRA-Det is listed in Table 1, including every stages of Swin Transformer backbone and two shared fully connected network layers in detection head. The original weight matrices dimensions, the low rank $r$ value and corresponding compressed parameter ratio $p$ is presented to concluded the trainable parameter efficiency of LoRA-Det. As the network depth increases, closer to the end of the network architecture, the proportion of trainable parameters in the LoRA module rapidly decreases relative to the original weight parameters. This confirms the proposed LoRA-Det's highly efficient trainable parameter compression capability.

Through experimental analysis, the proposed method, under the condition of only using a minimal proportion of the model's parameters for LoRA-based parameter-efficient fine-tuning and full fine-tuning for convolutional modules, significantly improves the pre-trained model's generalization performance by adopting new data. LoRA-Det achieves to approximate of the performance of full model fine-tuning, and in some cases, it is on par with full fine-tuning upper bound.

On the DOTAv1.0 dataset, the performance of the pre-trained model is measured as a mean Average Precision (mAP) of 73.74$\%$. Full fine-tuning of the model using 44.78 million parameters can elevate its performance to an upper bound of AP 77.84$\%$. However, fine-tuning only the backbone or only the Oriented R-CNN head results in AP of 50.2$\%$ and 67.8$\%$, respectively. Not only does this fail to enhance the model's detection capabilities, but it may also lead to model training collapse, deteriorating detection performance below the lower bound set by the pre-trained model. While utilizing LoRA modules to fine-tune backbone and full fine-tuning detection head can achieve satisfactory detection performance, it consumes 18.52 million trainable parameters, which is over three times the parameter quantity utilized by the proposed LoRA-Det.

\begin{table}[htbp]
\centering
\caption{The comparison results of the proposed method and other fine-tuning methods on DOTAv1.0 dataset. Evaluation by mAP and the number of trainable Parameters.}
\begin{tabular}{lccc}
\toprule
\textbf{Method}&Trainable& Trainable  & mAP  \\
&  Params(M) &Params Ratio($\%$)& (AP50)\\
\midrule 
Pre-trained  & 44.76 &100 & 73.74 \\
Full fine-tune & 44.76 & 100  & 77.84  \\
\midrule 
Full fine-tune backbone only& 30.25 &67.5 & 50.20  \\
Full fine-tune head only& 17.26 &38.5 & 67.80  \\
LoRA & 18.52 &40.9 & 77.03  \\
LoRA-Det & 4.92 & 10.9& 75.07 \\
LoRA-Det (hybrid) & 5.52 &12.3& 76.27 \\

\bottomrule
\end{tabular}
\end{table}

In contrast, our proposed method, integrating LoRA for parameter-efficient fine-tuning with full fine-tuning on convolutional layers, achieves an AP of 76.27$\%$ by training with just 5.52 million parameters, or 12.3$\%$ of the model's parameters. This equates to 97.98$\%$ of the model's performance upper bound. As illustrated in Table 2, the model performance, computational requirements, and the volume of trainable parameters for various strategies are compared and listed.

\begin{table}[htbp]
\centering
\caption{The comparison results of the proposed method and other fine-tuning methods on HRSC2016 dataset. Evaluation by mAP and the number of trainable Parameters.}
\begin{tabular}{lcccccc}
\toprule
\textbf{Method}&Trainable& Trainable  & mAP  \\
&  Params(M) &Params Ratio($\%$)& (AP50)\\
\midrule 
Pre-trained  & 44.78 &100 & 59.0 \\
Full fine-tune & 44.78 & 100 & 71.9  \\
\midrule 
Full fine-tune backbone only& 30.25 &67.5 & 50.2  \\
Full fine-tune head only& 17.26 &38.5 & 67.5  \\
LoRA & 18.53 &40.9 & 69.3  \\
LoRA-Det & 4.92 & 10.9& 68.3 \\
LoRA-Det (hybrid) & 5.52 &12.3 & 71.9 \\

\bottomrule
\end{tabular}
\end{table}

For the HRSC2016 dataset, which features a smaller volume of data and imbalanced categories, conducting fine-grained object detection tasks on the 31 categories of ship targets presents significant challenges. The proposed method is capable of achieving performance nearly equivalent to that of full fine-tuning, as demonstrated in Table 3. The performance of other strategies is inferior, consistent with the experimental results from the DOTAv1.0 dataset.

\begin{figure*}[t]
\centering
    \begin{minipage}[t]{\linewidth}
	   \centering
            \small\rotatebox{90}{\hspace{9pt}Pre-trained}
	       \includegraphics[width=.115\linewidth]{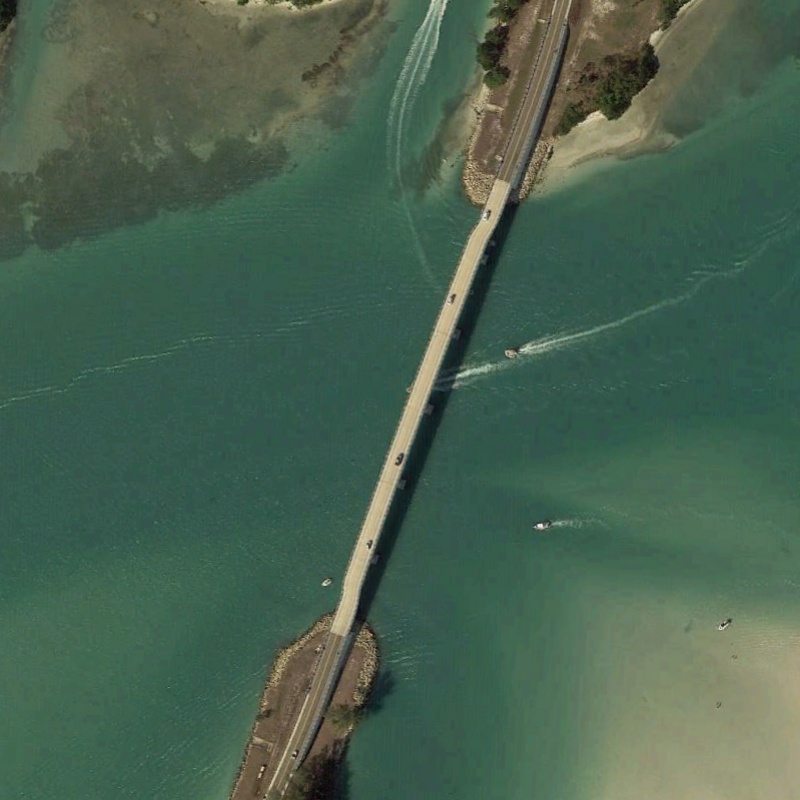}
	       \includegraphics[width=.115\linewidth]{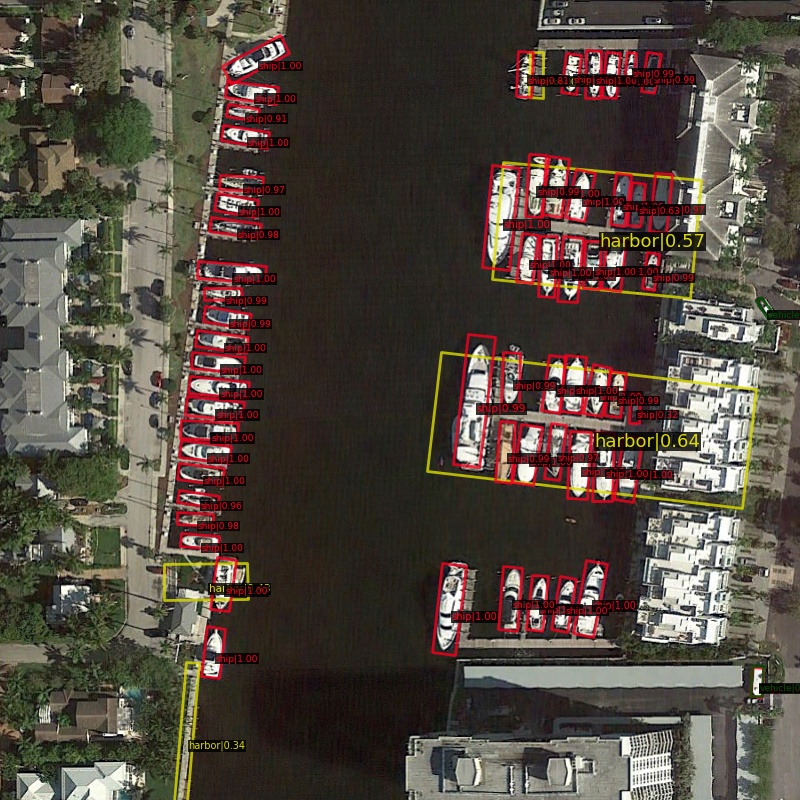}
              \includegraphics[width=.115\linewidth]{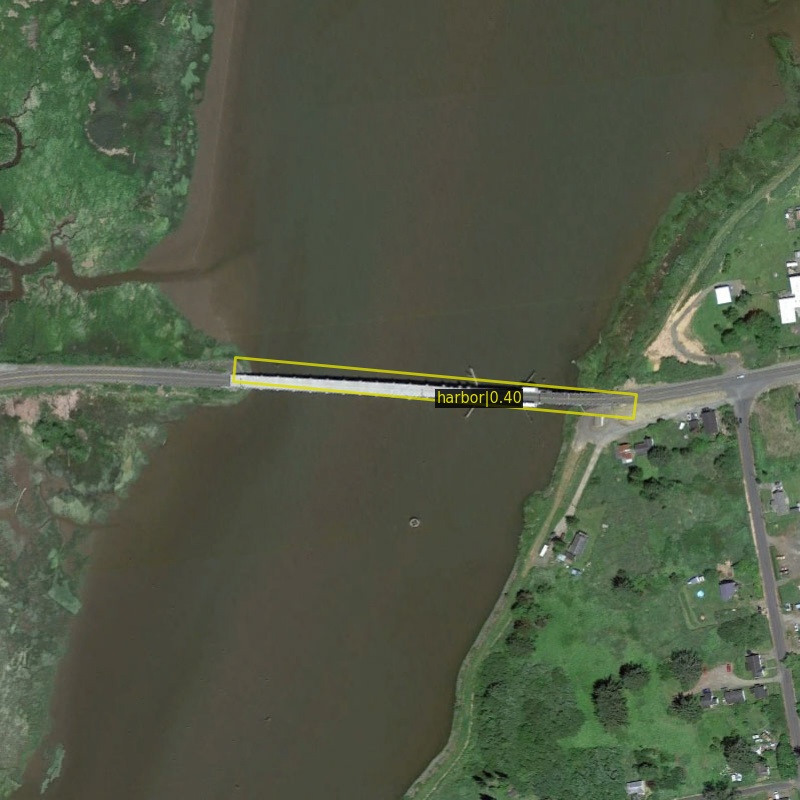}
              \includegraphics[width=.115\linewidth]{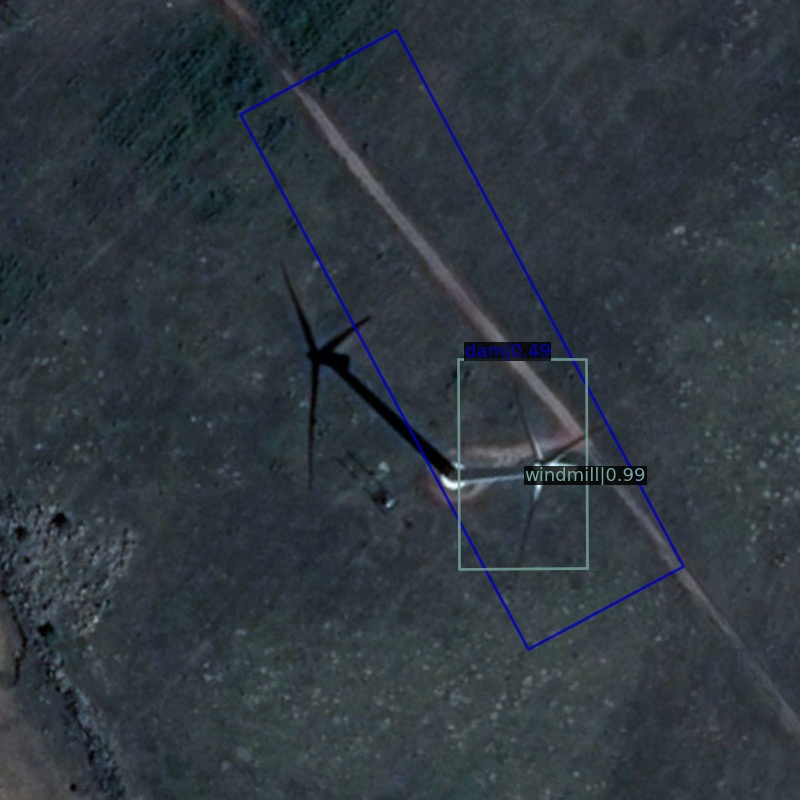}
              \includegraphics[width=.115\linewidth]{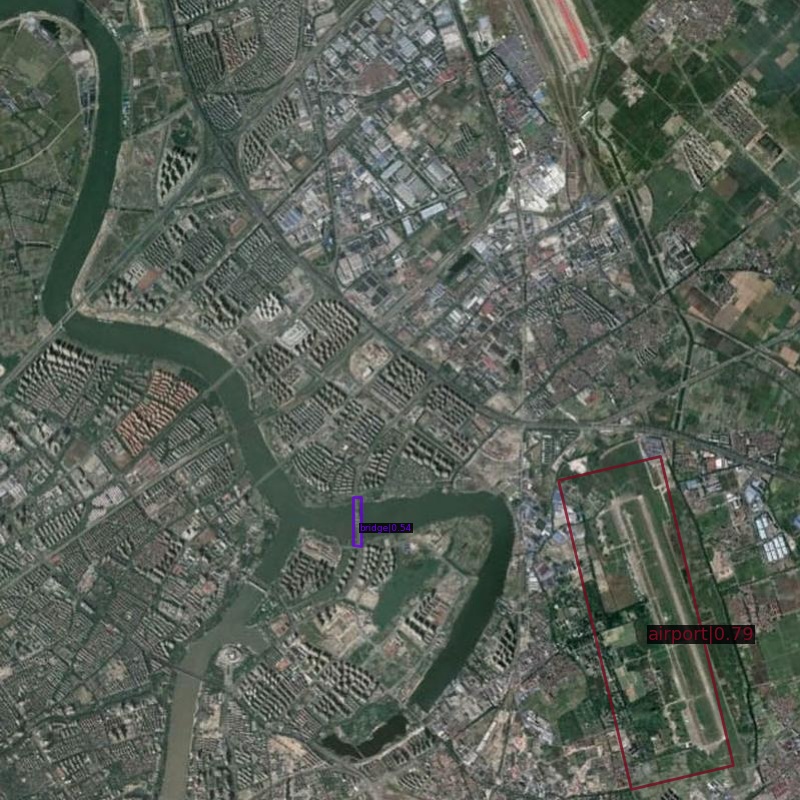}
              \includegraphics[width=.115\linewidth]{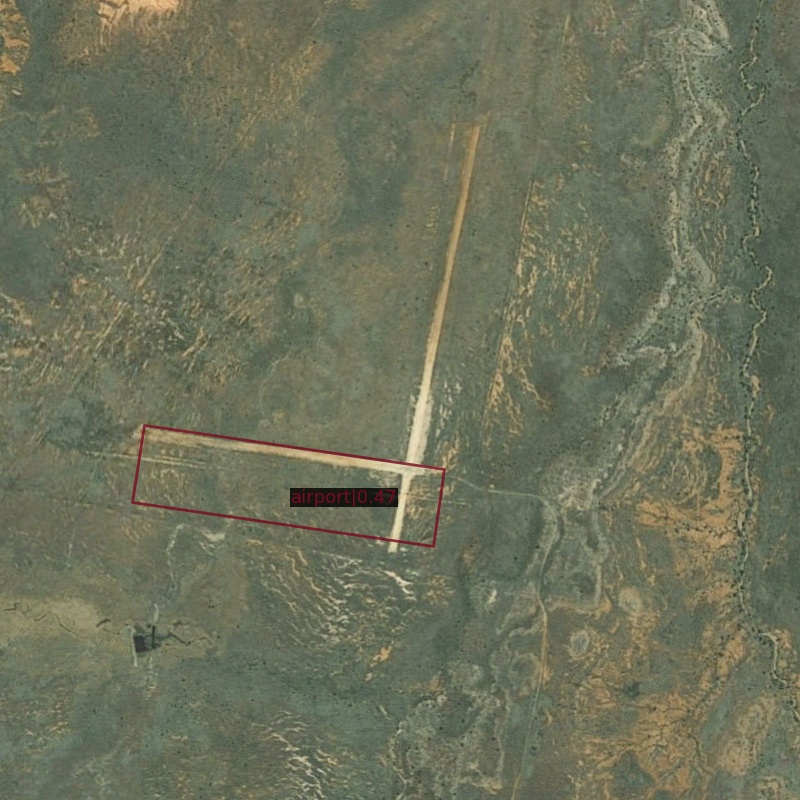}
              \includegraphics[width=.115\linewidth]{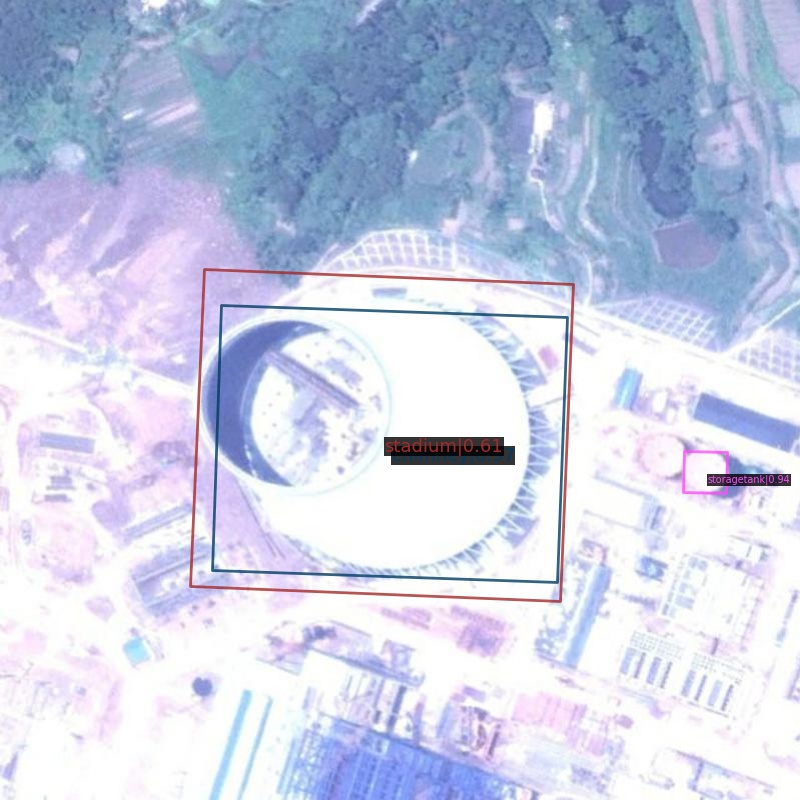}
              \includegraphics[width=.115\linewidth]{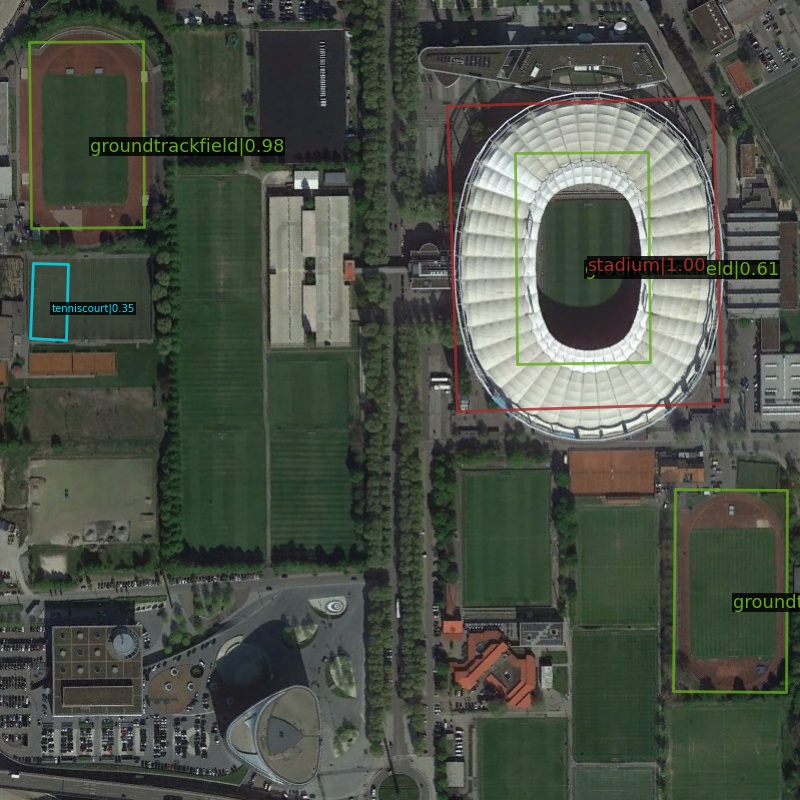}
            
	\end{minipage}\vspace{3pt}

    \begin{minipage}[t]{\linewidth}
	   \centering
            \small\rotatebox{90}{\hspace{5pt}Full fine-tune}
	       \includegraphics[width=.115\linewidth]{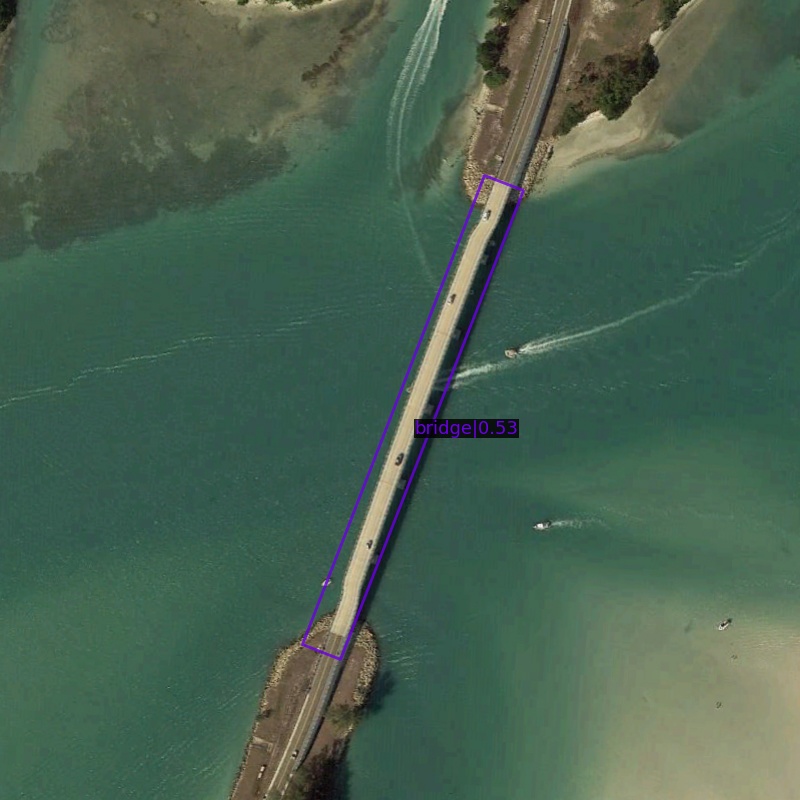}
	       \includegraphics[width=.115\linewidth]{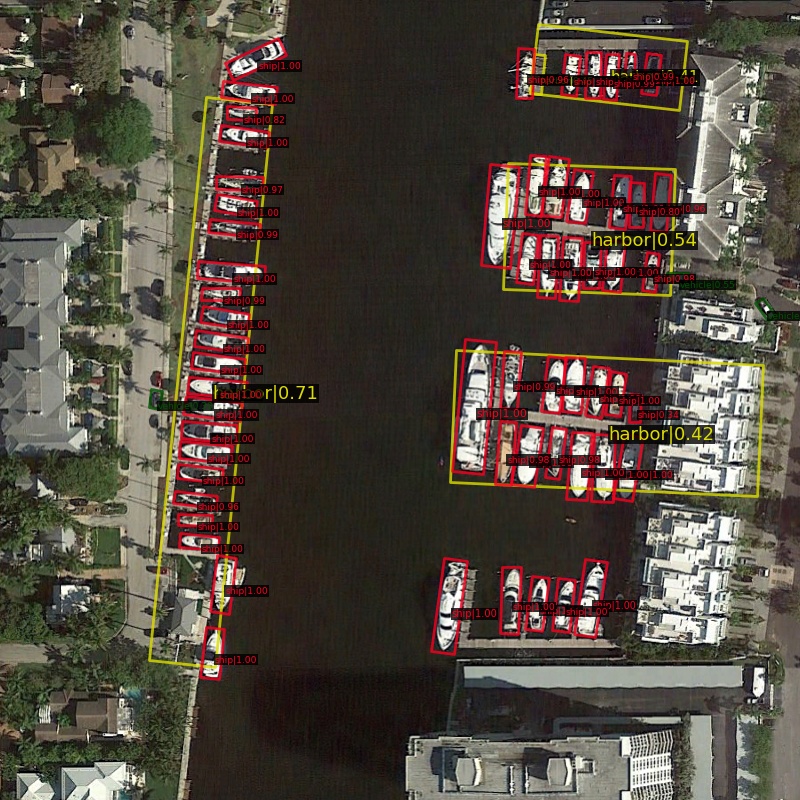}
              \includegraphics[width=.115\linewidth]{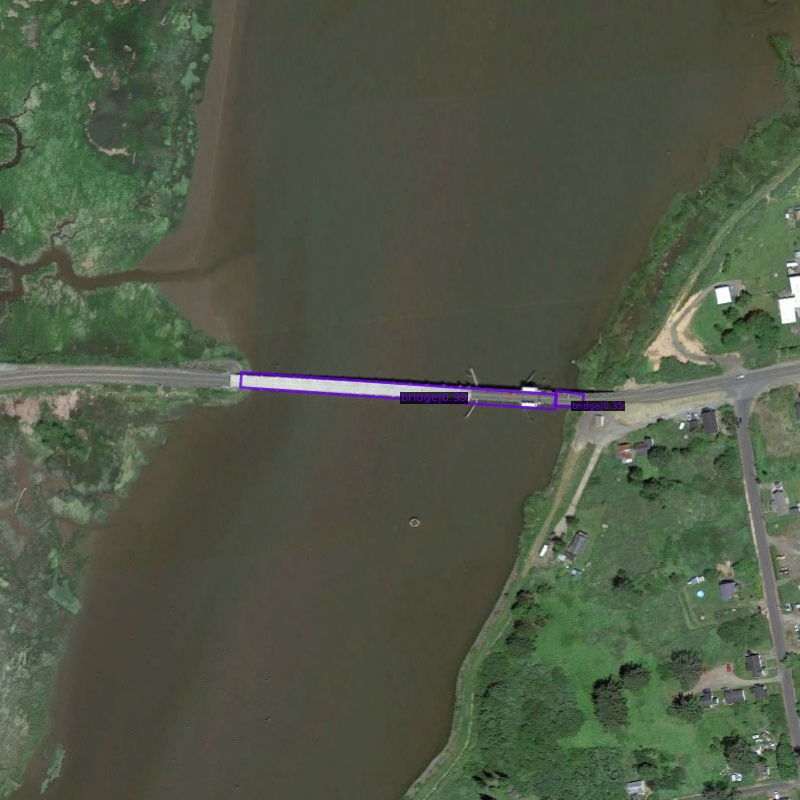}
              \includegraphics[width=.115\linewidth]{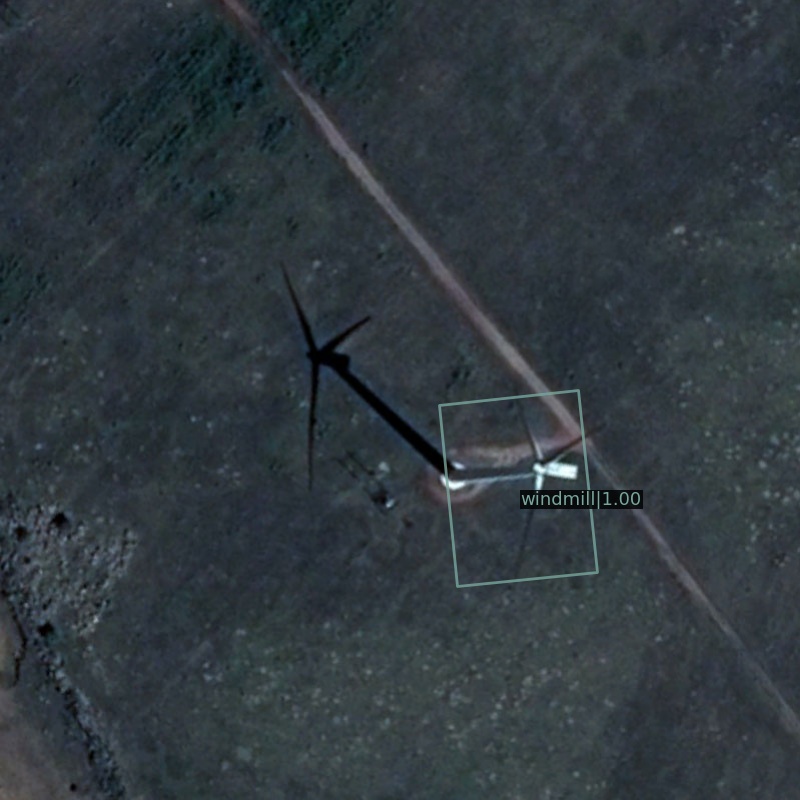}
              \includegraphics[width=.115\linewidth]{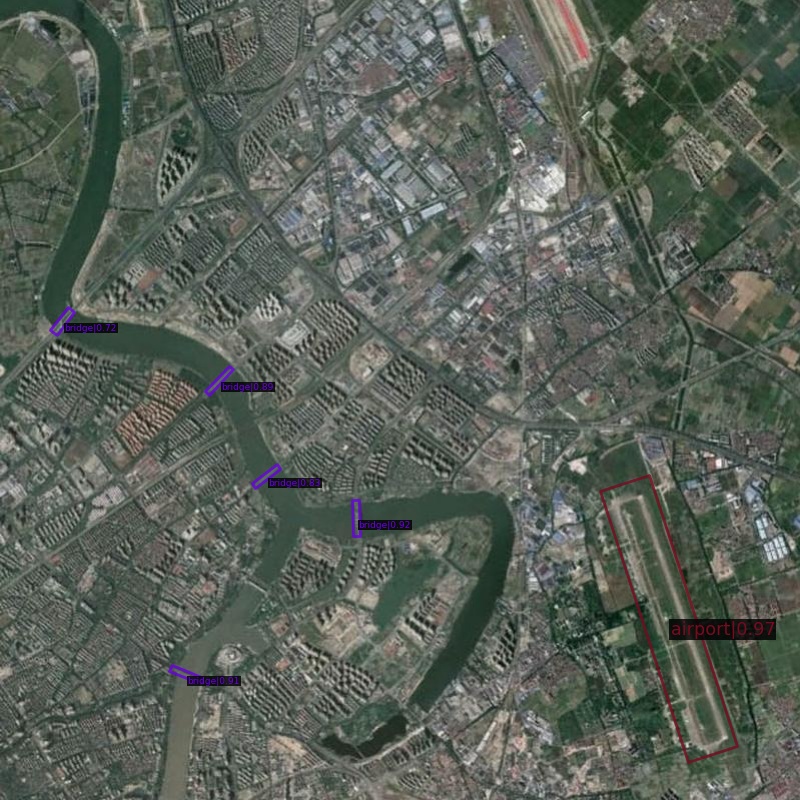}
              \includegraphics[width=.115\linewidth]{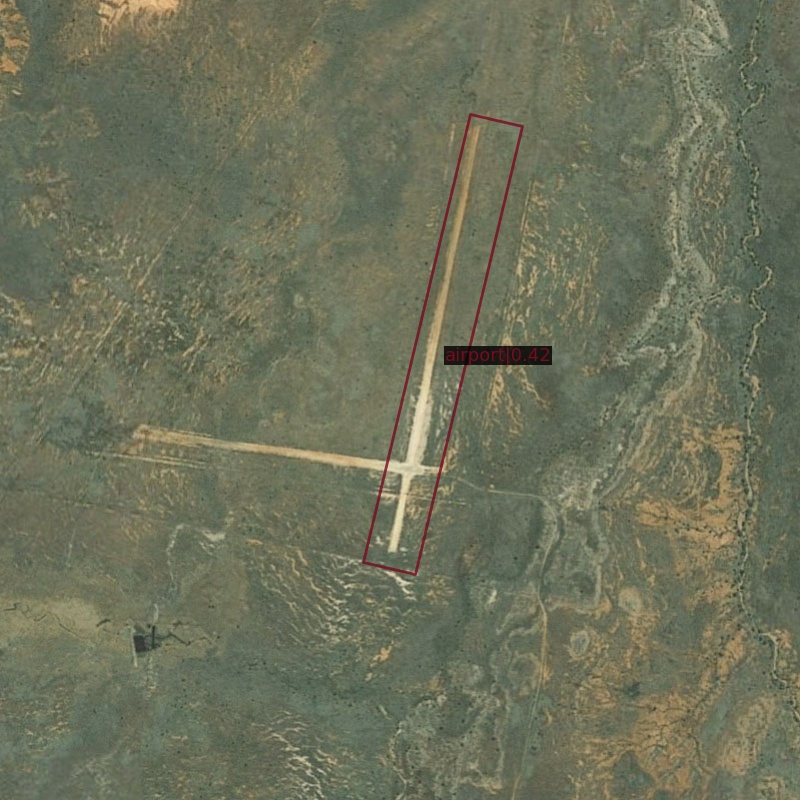}
              \includegraphics[width=.115\linewidth]{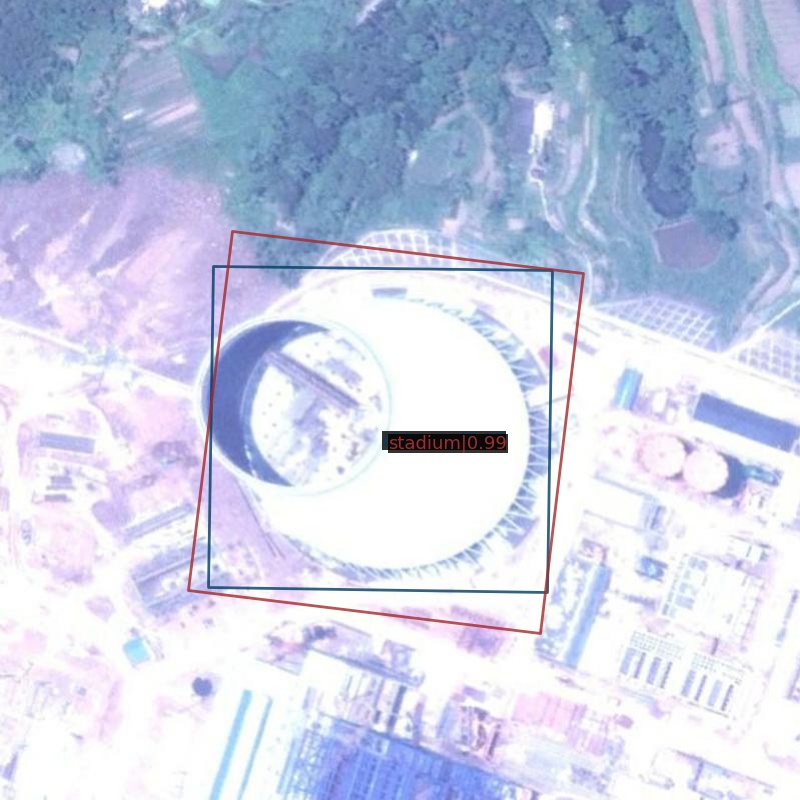}
              \includegraphics[width=.115\linewidth]{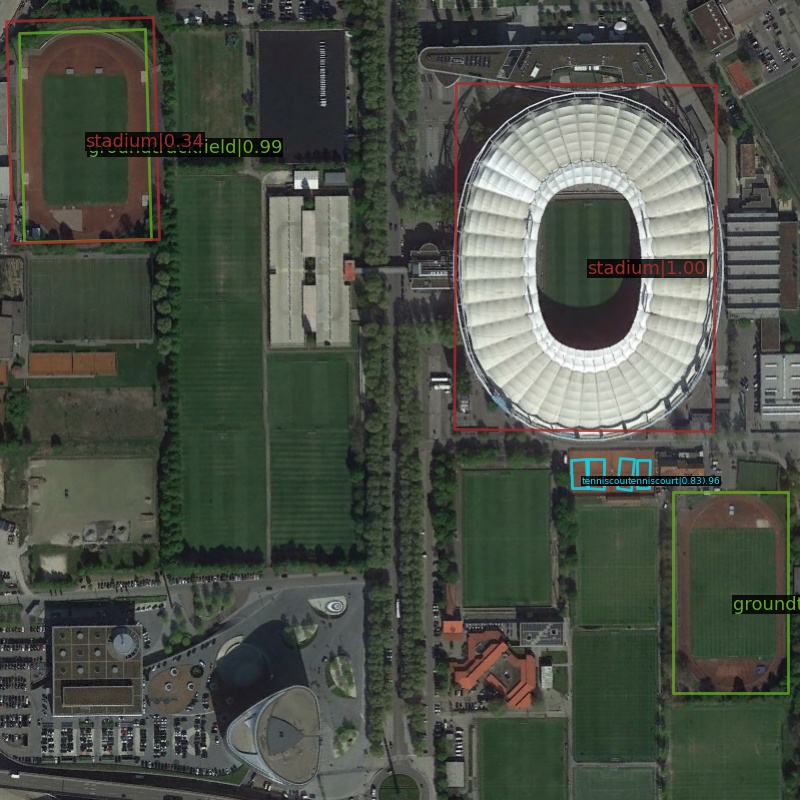}
	\end{minipage}\vspace{3pt}

    \begin{minipage}[c]{\linewidth}
	   \centering
            \small\rotatebox{90}{\hspace{8pt}LoRA-Det}
	       \includegraphics[width=.115\linewidth]{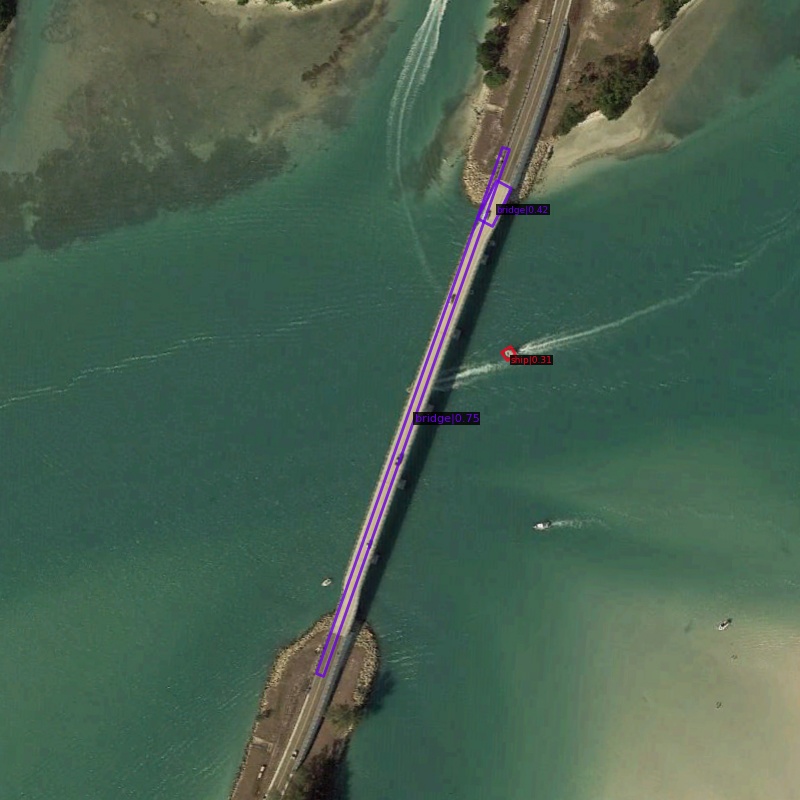}
	       \includegraphics[width=.115\linewidth]{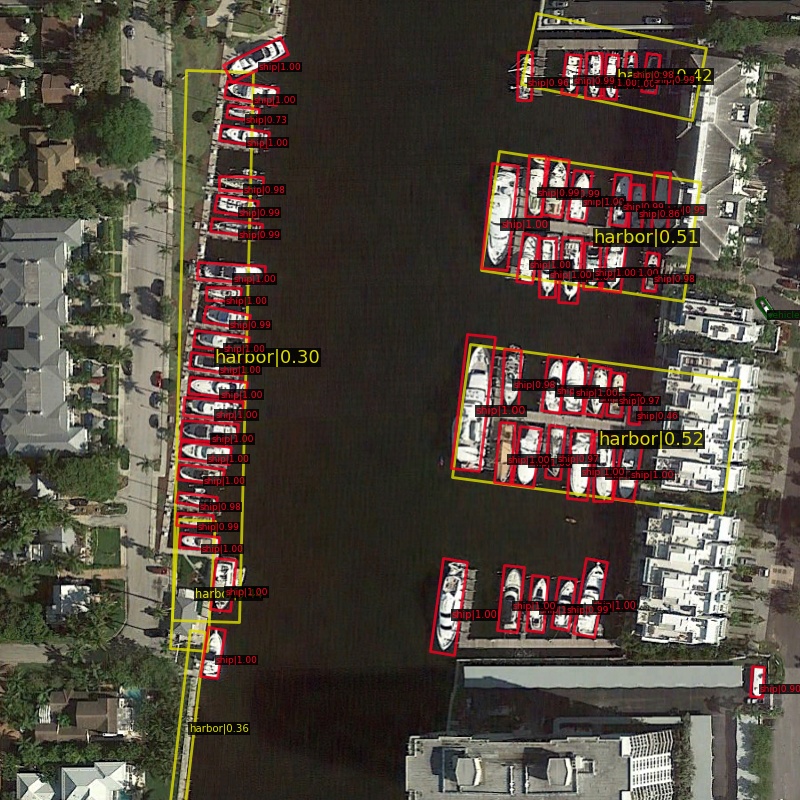}
              \includegraphics[width=.115\linewidth]{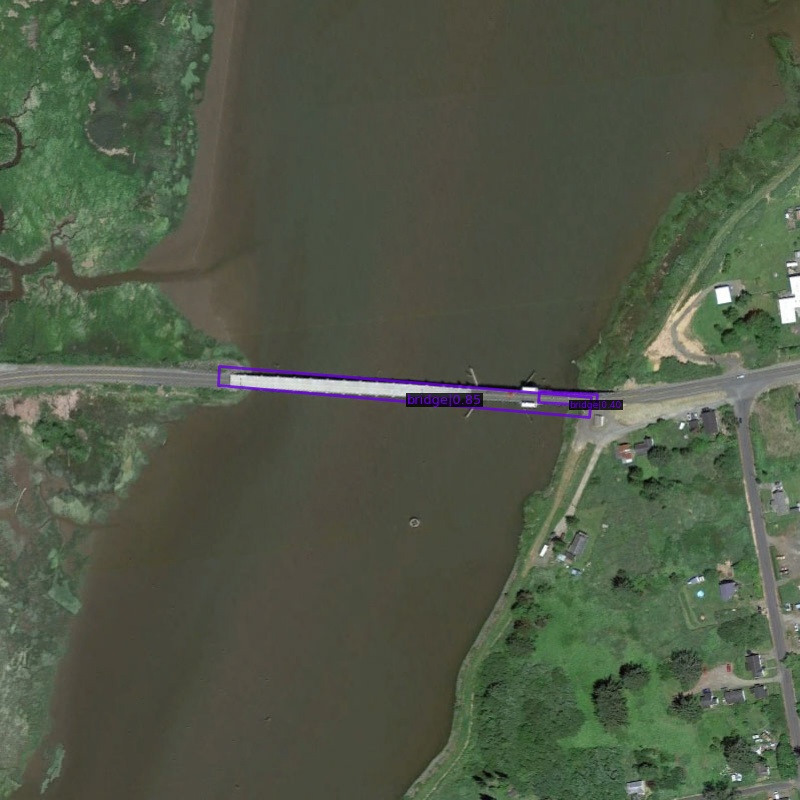}
              \includegraphics[width=.115\linewidth]{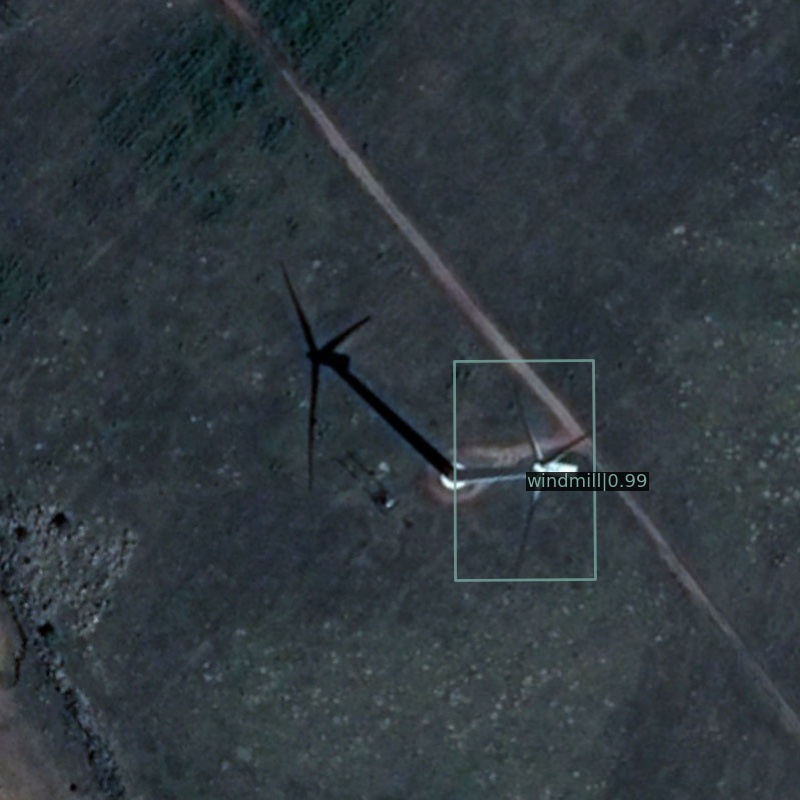}
              \includegraphics[width=.115\linewidth]{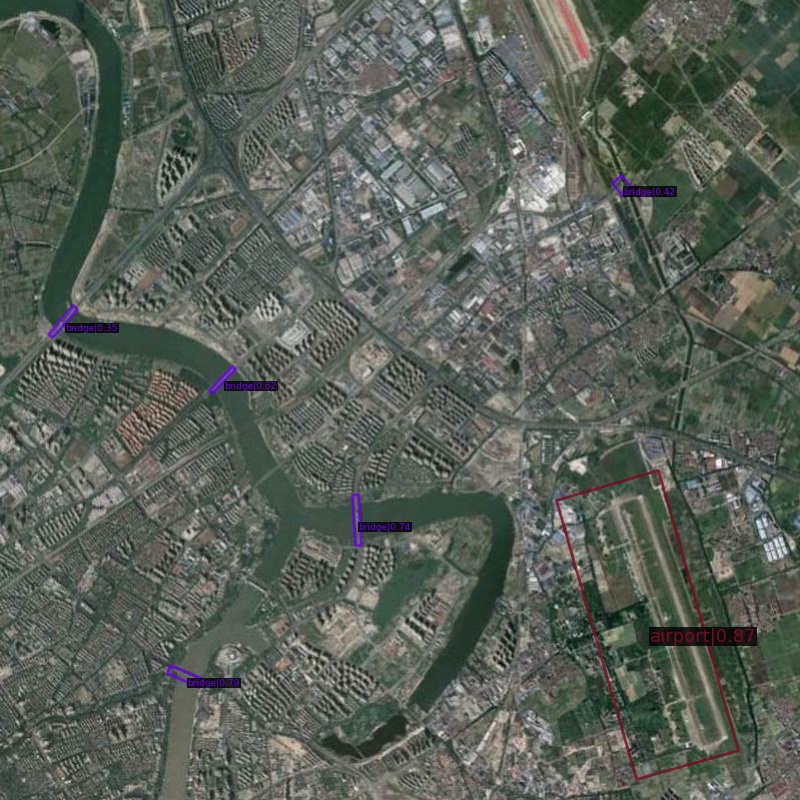}
              \includegraphics[width=.115\linewidth]{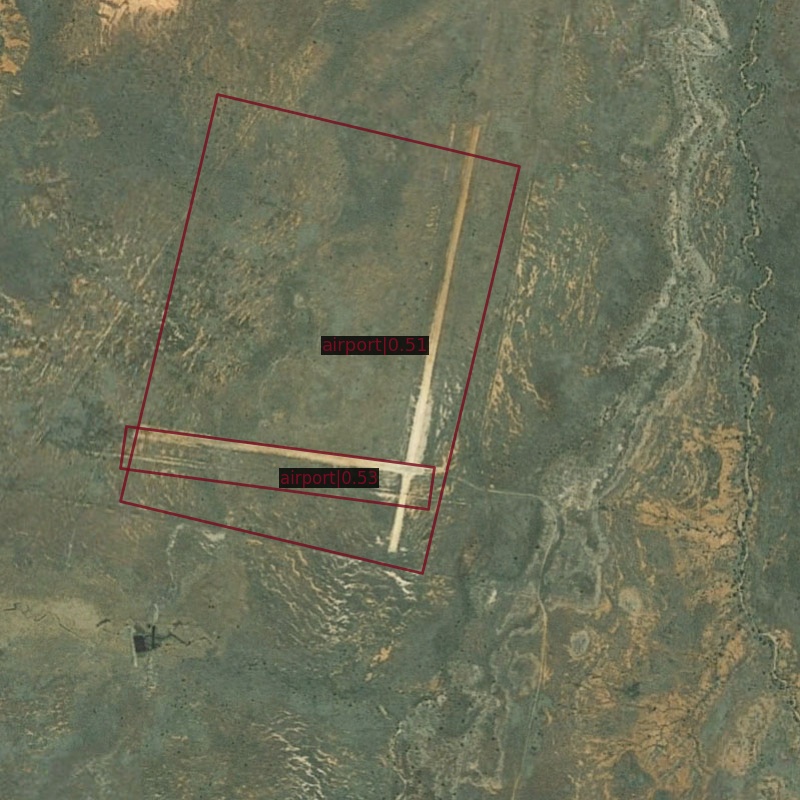}
              \includegraphics[width=.115\linewidth]{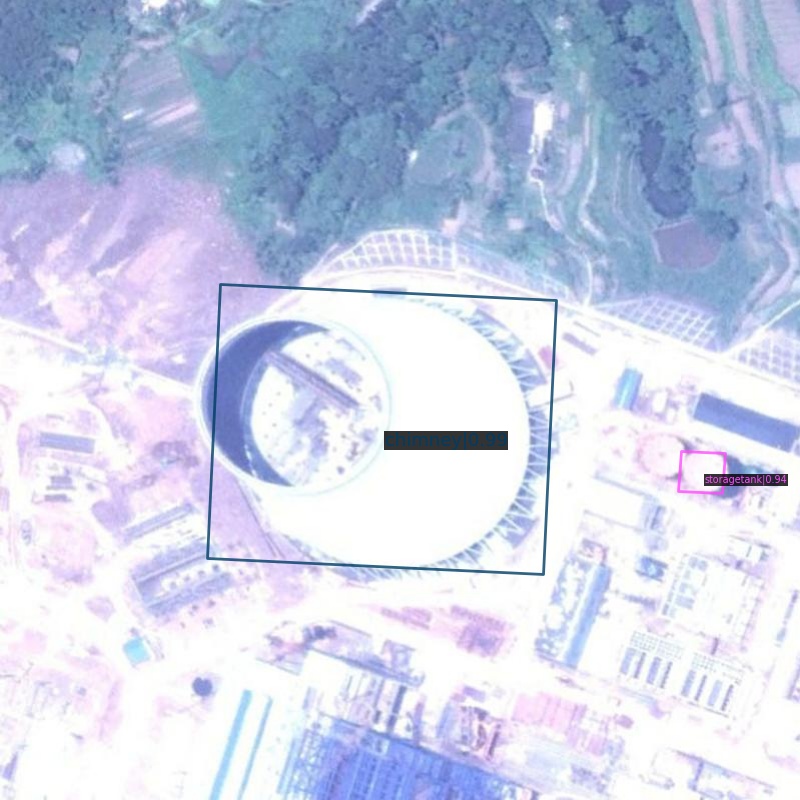}
              \includegraphics[width=.115\linewidth]{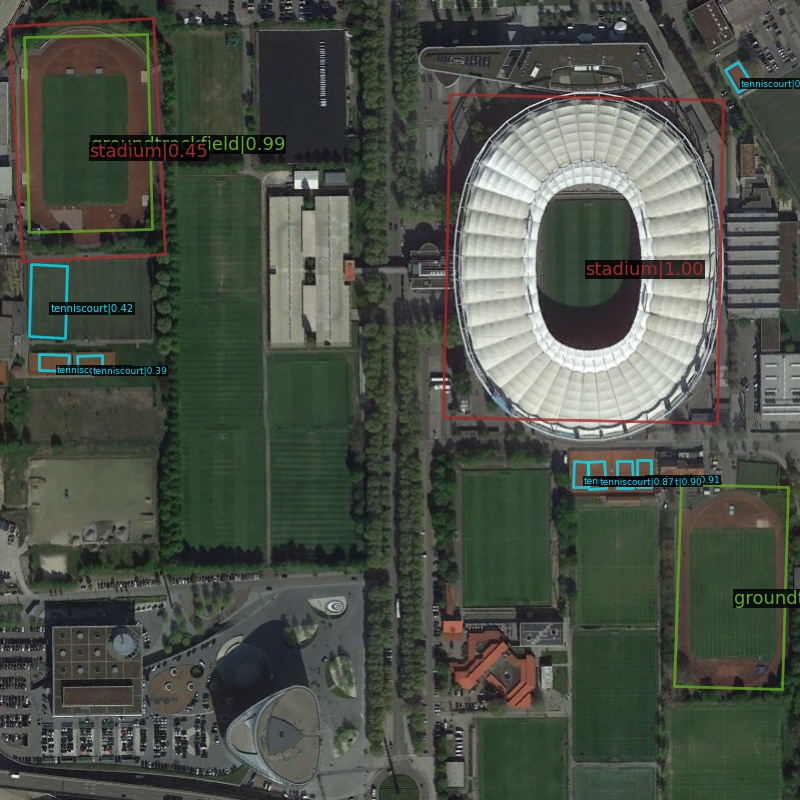}
	\end{minipage}\vspace{3pt}

    \begin{minipage}[t]{\linewidth}
	   \centering
            \small\rotatebox{90}{\hspace{3pt}Ground Truth}
	       \includegraphics[width=.115\linewidth]{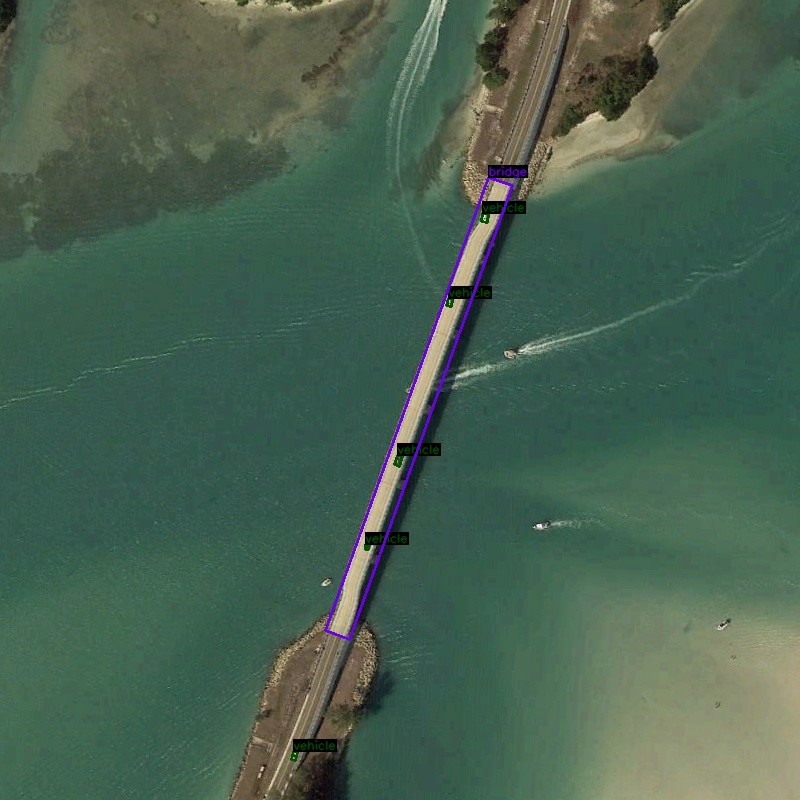}
	       \includegraphics[width=.115\linewidth]{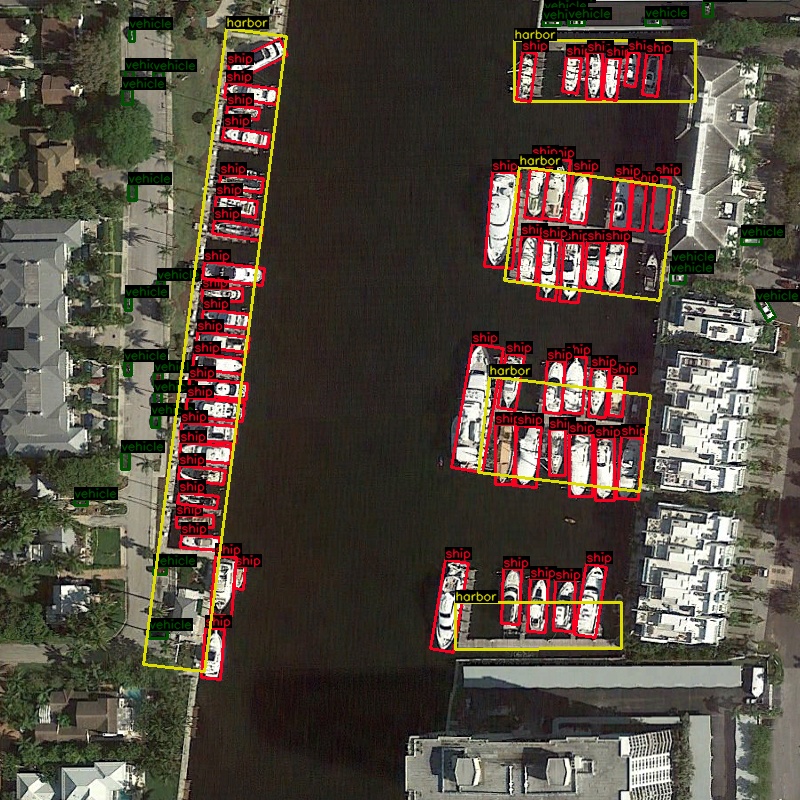}
              \includegraphics[width=.115\linewidth]{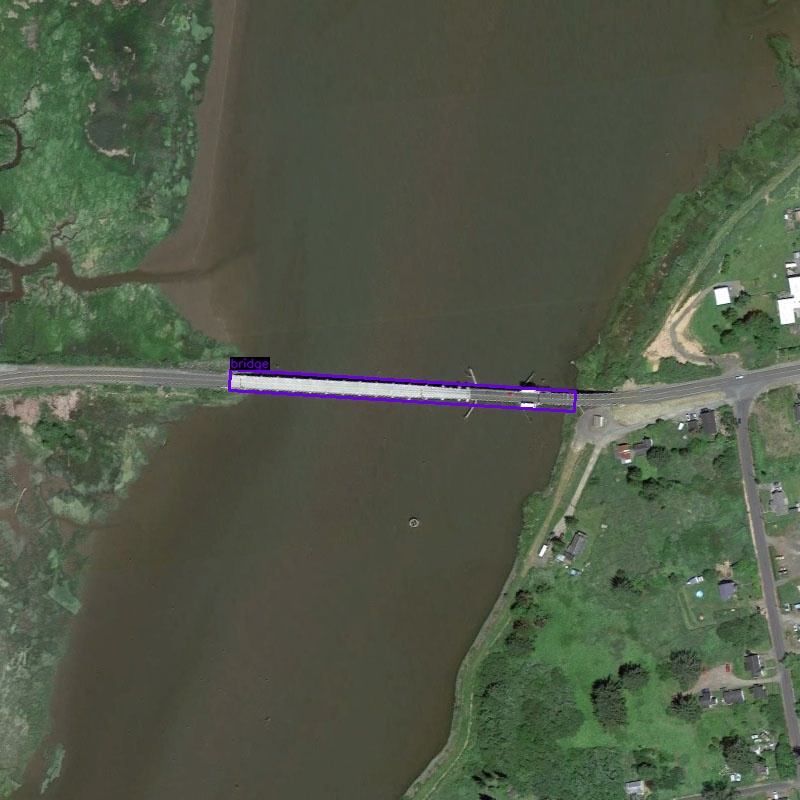}
              \includegraphics[width=.115\linewidth]{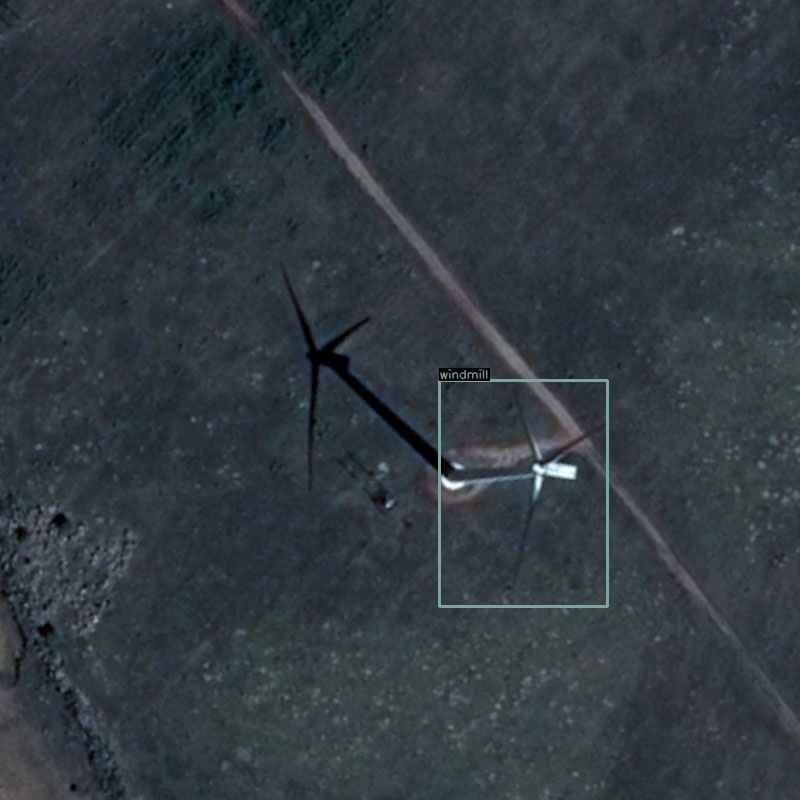}
              \includegraphics[width=.115\linewidth]{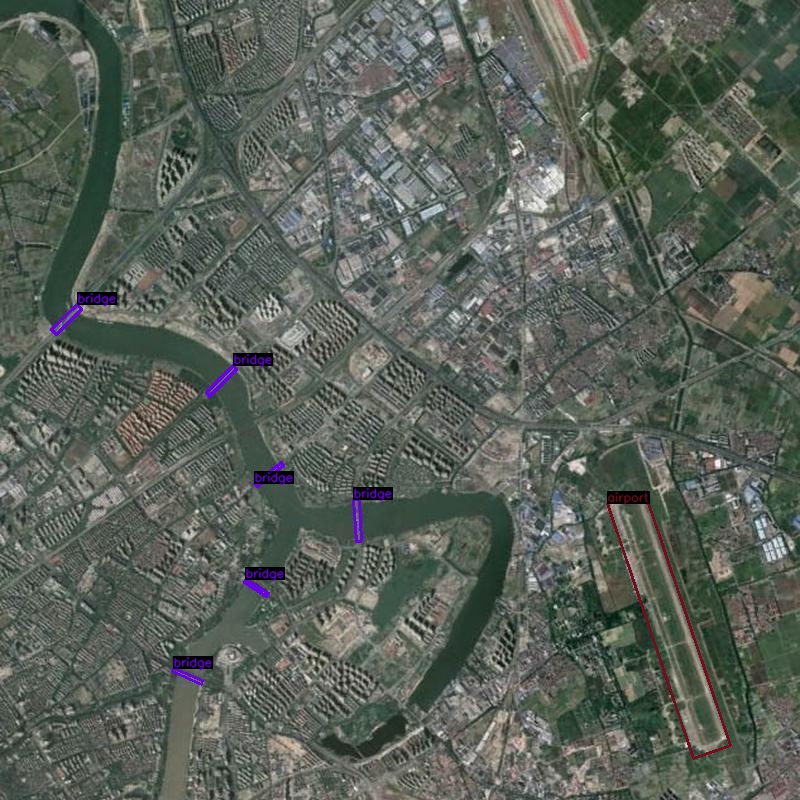}
              \includegraphics[width=.115\linewidth]{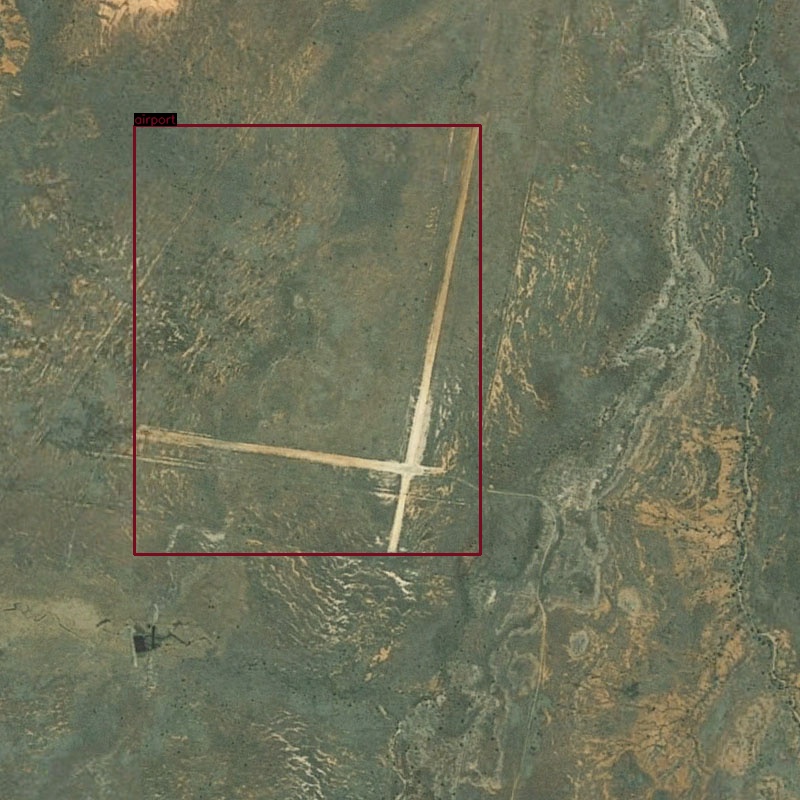}
              \includegraphics[width=.115\linewidth]{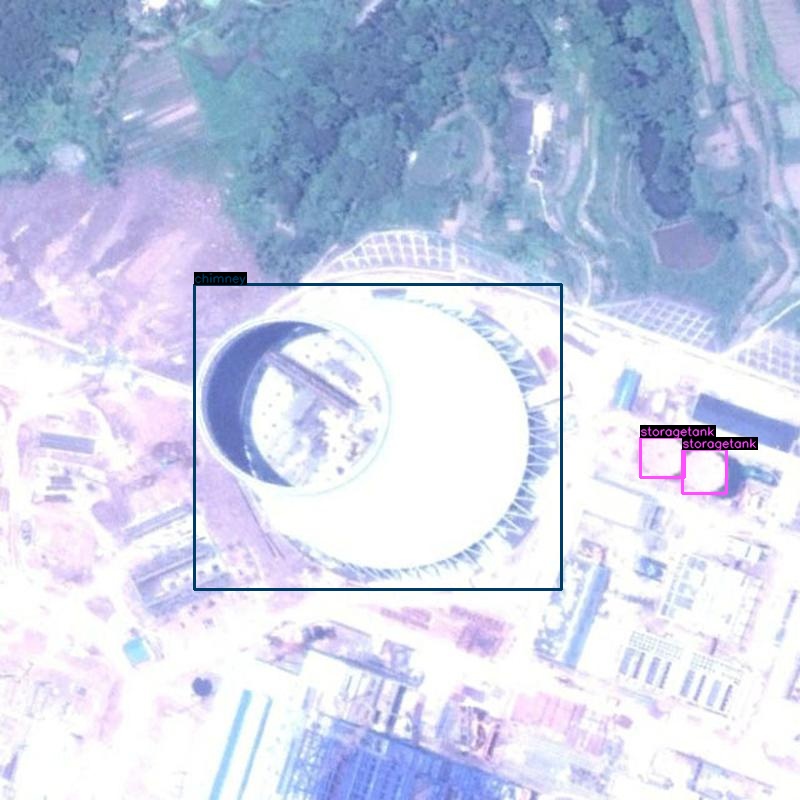}
              \includegraphics[width=.115\linewidth]{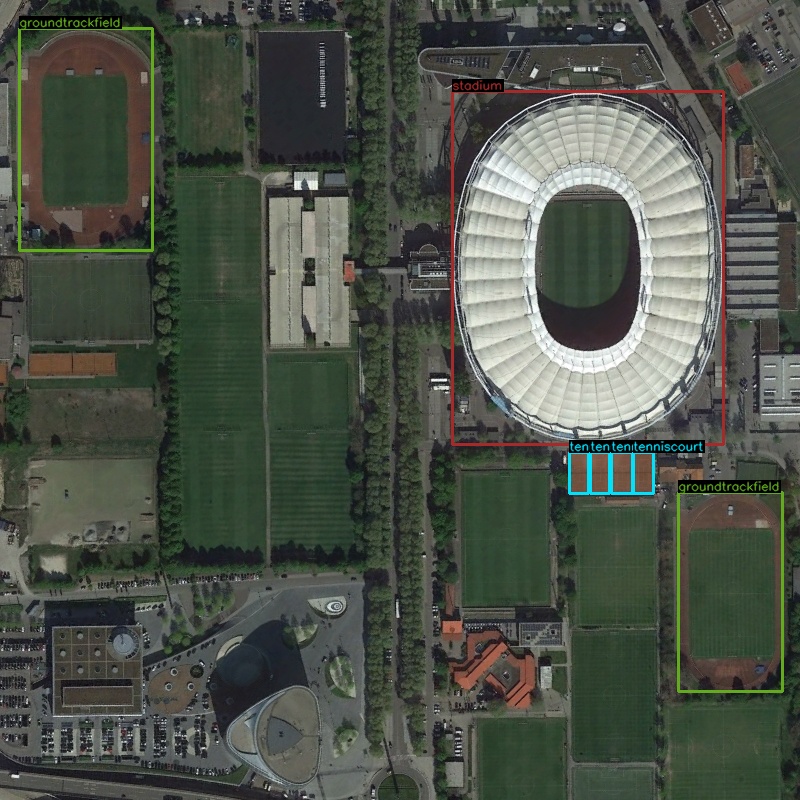}
	\end{minipage}\vspace{5pt}
 
\caption{Qualitative results of oriented object detection on DIOR-R testset. The compared visualizations between the proposed method and the pre-trained model, full fine-tuned model and ground truth. Diverse colors of bounding boxes represent different categories. The proposed method improves the issues of missing detection, false detection, and classification confusion in pre-trained models.}
\end{figure*}

\begin{table}[htbp]
\centering
\caption{The comparison results of the proposed method and other fine-tuning methods on DIOR-R dataset. Evaluation by mAP and the number of trainable Parameters.}
\begin{tabular}{lcccccc}
\toprule
\textbf{Method}&Trainable& Trainable  & mAP  \\
&  Params(M) &Params Ratio($\%$)& (AP50)\\
\midrule 
Pre-trained  & 44.77 &100  & 59.3 \\
Full fine-tune & 44.77 & 100 & 66.4  \\
\midrule 
Full fine-tune backbone only& 30.25 &67.5 & 33.2  \\
Full fine-tune head only& 17.26 &38.5 & 48.7  \\
LoRA & 18.52 &40.9 & 63.6 \\
LoRA-Det &  5.9 & 13.2 & 63.9 \\
LoRA-Det (hybrid) & 6.49 & 14.5 & 64.2 \\

\bottomrule
\end{tabular}
\end{table}

On the DIOR-R dataset, employing the proposed method to perform parameter-efficient fine-tuning on the pre-trained model with new data similarly yields significant benefits. A favorable balance between enhancing model performance and reducing the amount of parameters updated is achieved, consistent with the experimental conclusions drawn from the first two datasets. For this dataset, we select the different low-rank value $r$ in detection head, as 128 and 64 for the first and second shared fully connected layers. As listed in Table 4, The combination of LoRA-based parameter-efficient fine-tuning and full fine-tuning of certain convolutional modules of the detection algorithm reaches 96.75$\%$ of the full fine-tuning performance while using only 14.5$\%$ of the trainable parameters, and the Table 5 presents the detailed results of mAP on every categories on DIOR-R dataset. Figure 7 presents detection visualization of LoRA-Det and comparison with pre-trained model, full fine-tuned model and ground truth. LoRA-Det approaches the performance of fully fine-tuned models while also addressing issues of missing detection, false positives, and classification confusion in the pre-trained model.

\begin{table*}[t]
\renewcommand\arraystretch{1.5}
\vspace{-0.1cm}
\small
\setlength\tabcolsep{4.5pt}
\caption{Comparison with other fine-tuning methods on the DIOR-R dataset for every categories.}
\resizebox{\textwidth}{!}{
\begin{tabular}{lccccccccccccccccccccccc}
\hline
Methods     & mAP & APL    & APO    & BF    & BC    & BR    & CH       & ETS    & ESA   & DAM      & GF   & GTF   & HA& OP   & SH   & STA   & STO   & TC   & TS   & VE   & WM   \\ \hline
Pre-trained   & 59.3 & 71.1   & 31.5   & 71.2   & 81.3      & 34.6    & 72.7  & 59.4 & 51.5   & 23.3   & 69.1   & 78.1 & 34.4   & 50.0  & 81.0   & 71.2    &70.9   & 81.3   & 52.9   & 42.8    &   56.7\\
Full fine-tune   & 66.4 & 71.3  & 46.8  & 79.3   & 87.8   &43.8   & 78.9   & 83.6   & 69.9   & 29.4    & 77.7   & 83.4   & 42.7& 59.1   & 81.1   & 81.9   & 62.5   & 81.5   & 57.6   & 43.5   & 65.4   \\  \hline
LoRA & 63.6 & 71.9 & 36.9   & 78.5 & 86.3  & 39.0  & 72.6 &78.2 &66.0 &25.3 &68.9 &83.0 &34.4 &53.0 &81.1 &81.2 &70.7 &81.4 &52.0 &46.6 &64.6  \\
LoRA-Det & 63.9 &  71.5 &39.2 &78.6&81.4 &39.0 &72.7 &76.0 &64.9 &25.6 &73.2 &82.5 &39.5 &53.9 &81.1 &80.7 &71.1 &81.5 &54.1 &47.4 &64.3   \\ 
LoRA-Det (hybrid)  & 64.2 &71.3 &38.1 &78.5 &86.1 &39.0 &78.5 &76.4 &64.9 &27.0 &73.4 &81.8 &39.0 &54.3 &81.1 &80.5 &71.1 &81.5 &53.8 &43.3 &64.6 \\ \hline
\end{tabular}
}
\label{table:dior_comparison}
\end{table*}

\begin{figure*}[tbhp]
\centering
    \begin{minipage}[t]{\linewidth}
	   \centering
	       \includegraphics[width=.32\linewidth]{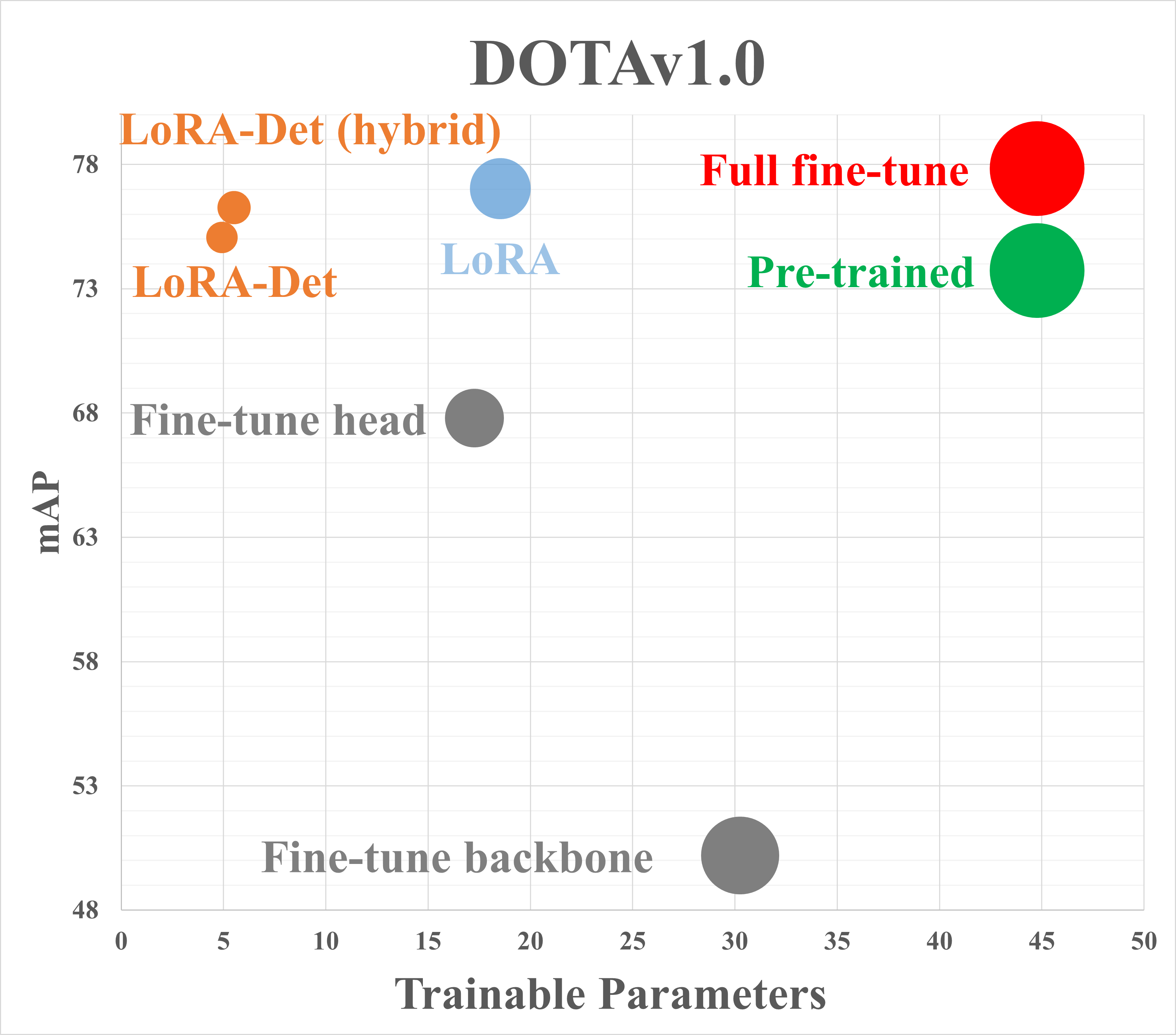}
	       \includegraphics[width=.32\linewidth]{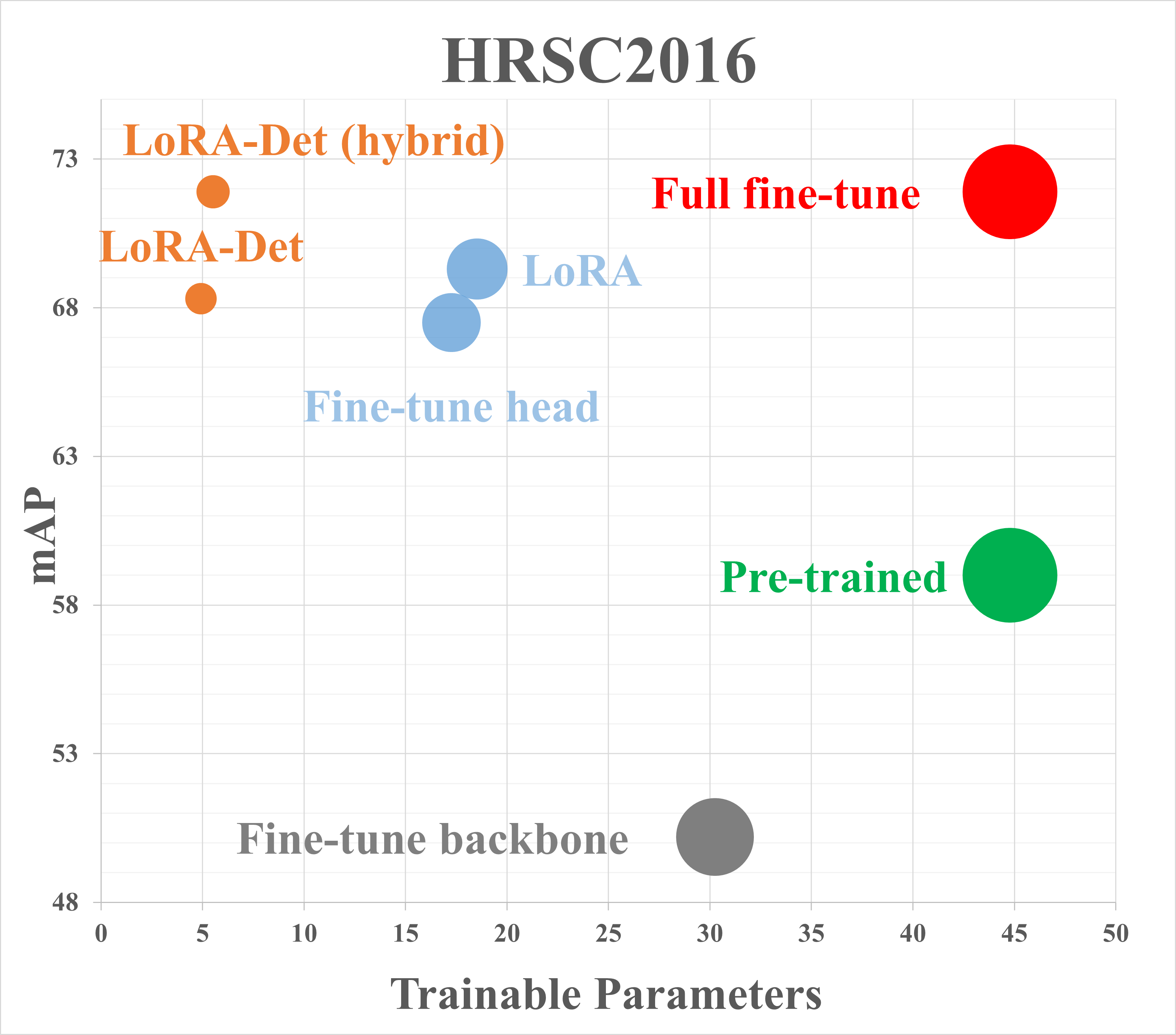}
              \includegraphics[width=.32\linewidth]{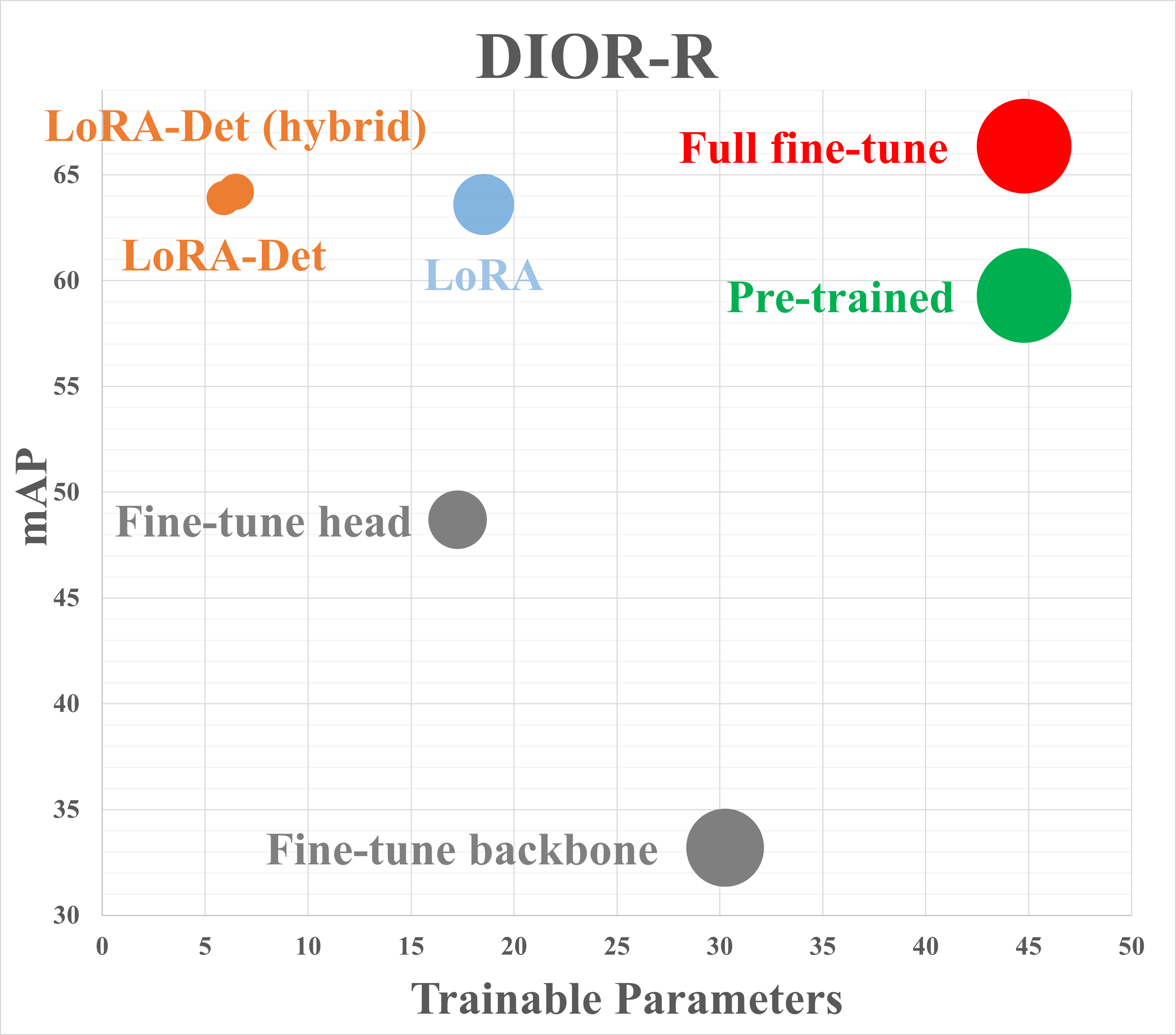}
	\end{minipage}\vspace{3pt}

\caption{Trainable parameter quantity vs. oriented object detection performance mAP of the proposed LoRA-Det and other fine-tuning methods listed in the previous tables. Green bubbles represent the lower bound of pre-trained models, red bubbles are upper bound of full fine-tuning, gray bubbles mean the performance lower than pre-trained models. The size of the bubbles indicates the volume of trainable parameters.}
\end{figure*}

Figure 8 illustrates the comparison of model performance and the quantity of trainable parameters between the proposed LoRA-Det and other fine-tuning approaches across three remote sensing image datasets. The horizontal axis represents the amount of trainable parameters, while the vertical axis denotes the model's object detection performance, measured in mAP \cite{mAP}. The size of the bubbles indicates the volume of trainable parameters; hence, bubbles located towards the upper left corner of the image, which are smaller in area, signify samples with superior overall performance. 

The visualization results from experiments on the DOTAv1.0, HRSC2016, and DIOR-R datasets lead to a consistent conclusion: LoRA-Det, even with a significantly reduced amount of trainable parameters, still achieves competitive detection performance, managing to update far fewer parameters while approaching the performance of models fully fine-tuned. Similarly, the experiments validate that even with a larger number of trainable parameters, inappropriate fine-tuning methods fail to achieve optimal detection results.

\subsection{Comparison with Lightweight State-of-the-Art Method}
To validate the applicability of the proposed algorithm, it is compared with the full fine-tuning performance of lightweight detection algorithms. The comparison involves the parameter-efficient fine-tuning techniques of the normal-size model versus the full fine-tuning of lightweight detection models under similar amounts of trainable parameters, i.e., within the constraints of satellite model update parameter limitations. The experiment compares the full fine-tuning performance of the lightweight oriented detection model RTMDet \cite{rtmdet} across different parameter specifications on the DOTAv1.0 dataset against the LoRA-Det model, which only fine-tuned 12.4$\%$ of all parameters. Both models had comparable trainable parameter volumes. The training configurations of RTMDet are consistent with the original specifications on single-scale level in \cite{rtmdet}. The data augmentation strategy is only involved random flipping, which is same as LoRA-Det.

\begin{table*}[t]
\renewcommand\arraystretch{1.5}
\centering
%\vspace{-3pt}
\vspace{-0.1cm}
\small
\setlength\tabcolsep{4.5pt}
\caption{Comparison of LoRA-Det with lightweight detector RTMDet on the number of trainable parameters, and mAP accuracy on DOTAv1.0 testset. The AP of each category is listed.}\label{tab:rtmdet-dota}
\resizebox{\textwidth}{!}{
\begin{tabular}{lcccccccccccccccccc}
\toprule
% \hline
Method & Trainable Params(M)&  mAP &  PL &  BD &  BR &  GTF &  SV &  LV &  SH &  TC &  BC &  ST &  SBF &  RA &  HA &  SP &  HC \\ 
\midrule 
RTMDet-tiny  & 4.88  & 73.88 & 88.63 & 76.14 & 48.99 & 66.82 & 80.57 & 80.66 & 88.28 & 90.83 & 85.00 & 87.01 & 54.81 & 62.44 & 68.15 & 75.63 & 54.30 \\
RTMDet-small  & 8.86  & 75.24 & 88.71 & 79.89 & 50.45 & 69.82 & 80.19 & 81.73 & 88.68 & 90.86 & 82.78 & 87.44 & 55.38 & 65.12 & 72.80 & 80.56 & 54.22 \\
\midrule 
LoRA-Det  & 4.92  & 75.07 & 89.38 & 79.56 & 51.59 & 72.10 & 77.48 & 83.13 & 87.80  & 90.88 & 84.52 & 85.42 & 60.44 & 65.35 & 66.24 & 67.28 & 64.82 \\
LoRA-Det (hybrid)  &  5.52 & 76.27 & 89.25 & 81.33 & 53.03 & 75.53 & 77.91 & 83.05 &87.89 & 90.89 & 85.77 & 85.10 & 63.42 & 67.73 & 73.33 & 68.35 &61.44 \\
\bottomrule
\end{tabular}
}
\end{table*}

As shown in Table 6, the comparison of detection performance (measured by mAP) and trainable parameter volumes between LoRA-Det and RTMDet (tiny and small size) demonstrates that LoRA-Det, with a similar volume of trainable parameters under the PEFT approach, surpasses the lightweight detection models RTMDet-tiny, and even exceeds the performance of RTMDet-small, which has a higher volume of trainable parameters. This suggests that the proposed LoRA-Det, leveraging a larger overall model parameter scale combined with PEFT technology, offers better model performance and generalizability compared to lightweight models. It is well-suited for deployment in satellite communication application scenarios. LoRA-Det presents great potentiality on improving model performance of spaceborne detectors by parameter-efficient fine-tuning on continuous data under the data update limitation and enough inference resources.

\subsection{Analysis of Low Rank Approximation}

To select an appropriate low-rank value $r$, it is necessary to analyze whether the parameter weight matrix of the pre-trained model has a relatively low intrinsic dimension, or is close to full rank. Utilizing the proposed low rank approximation method based on Singular Value Decomposition matrix decomposition, our focus is on analyzing the low rank approximation experiments of the trainable parameter matrices in the proposed LoRA-Det method. Specifically, the parameter matrices for matrix decomposition approximation originate from the pre-trained weights of the Oriented RCNN model on the DOTAv1.0 dataset. 

Given that each layer’s trainable parameter matrix dimensions differ, as shown in Table 1, the maximum dimension for the decomposition rank of the low-rank approximation experiment is determined by the matrix's smallest dimension, denoted as \(rank_{max} = \min(d, k)\). For example, in the LoRA modules of Swin Transformer backbone stages 1, 2, 3, and 4, the maximum values for the low-rank estimation decomposition dimensions $r$ are respectively 96, 192, 384, 768, consistent with the dimensions of the query, key, value parameter matrices in the multi-head attention of that stage. Similarly, for the two shared fully connected layers of the detection head, the maximum value for the low-rank approximation decomposition dimension $r$ is 1024, equal to \(\min(1024, 12544)\).

\begin{figure}[!ht]
\centering
\includegraphics[width=3.5in]{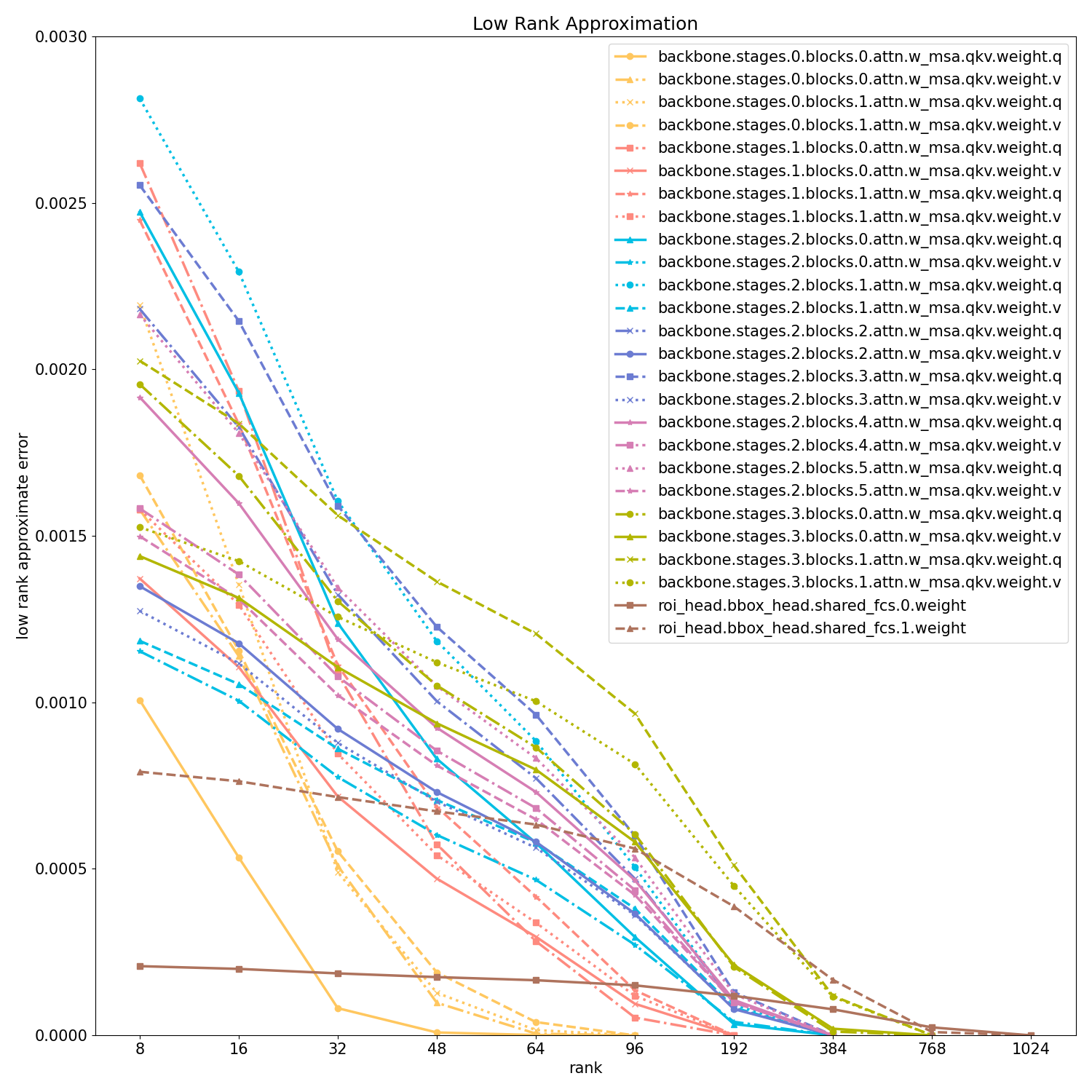}
\caption{Low rank approximation results of each trainable parameter matrices in LoRA module of LoRA-Det.}
\label{fig1}
\end{figure}

Figure 9 displays the results of the low rank approximation for the trainable parameter matrices of the LoRA module in LoRA-Det. The trainable matrices in the backbone show a relatively high approximation error, which gradually decreases as the approximation rank increases. This indicates that the parameter matrices in the backbone of the pre-trained weights have a large intrinsic dimension, even approaching full rank. In contrast, for the two layers of shared fully connected networks in the detection head, despite the matrices' large original dimensions ((1024, 12544) and (1024, 1024)), the low-rank approximation error remains small as the rank $r$ increases. This suggests that these parameter matrices have a small intrinsic dimension, allowing for accurate approximation with lower-rank decomposed matrices.

\begin{figure}[htbp]
  \centering
  % 第一行第一个子图
  \begin{minipage}{0.485\linewidth}
    \centering
    \includegraphics[width=\linewidth]{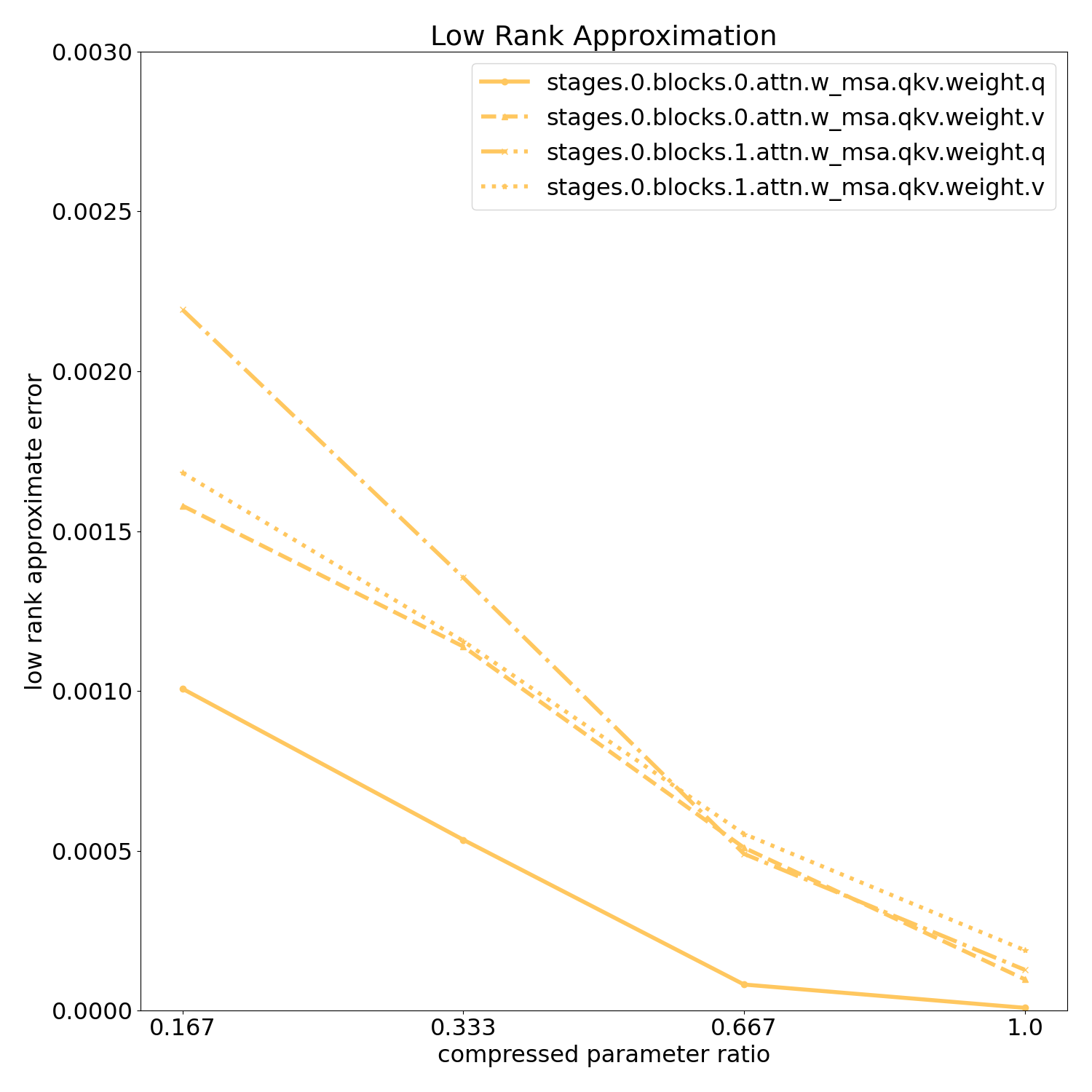}
    %\caption{backbone stage1}
   % \label{fig:sub1}
  \end{minipage}
  \hfill % 用于子图之间的间隙
  % 第一行第二个子图
  \begin{minipage}{0.485\linewidth}
    \centering
    \includegraphics[width=\linewidth]{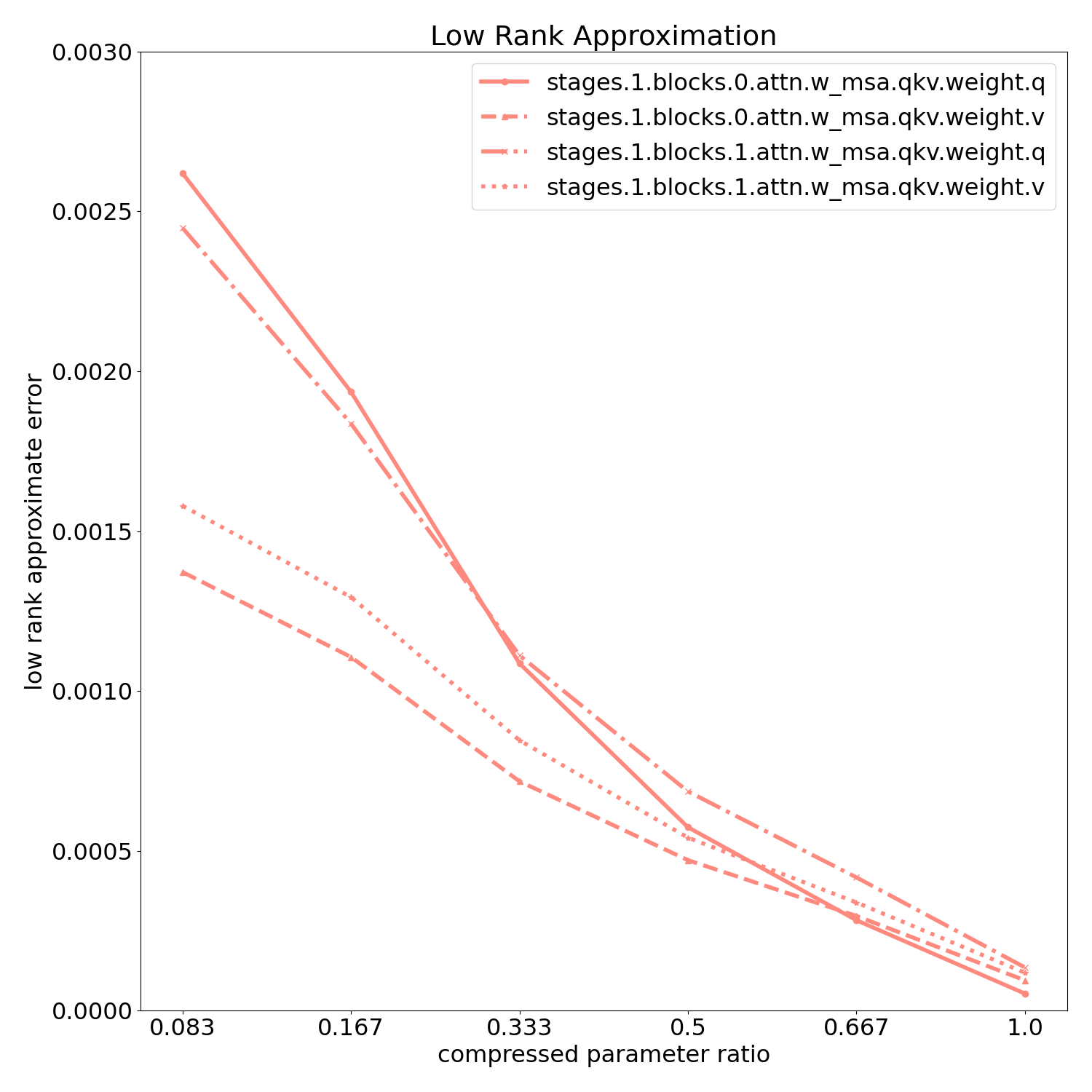}
   % \caption{backbone stage2}
    %\label{fig:sub2}
  \end{minipage}
  
  % 第二行第一个子图
  \begin{minipage}{0.485\linewidth}
    \centering
    \includegraphics[width=\linewidth]{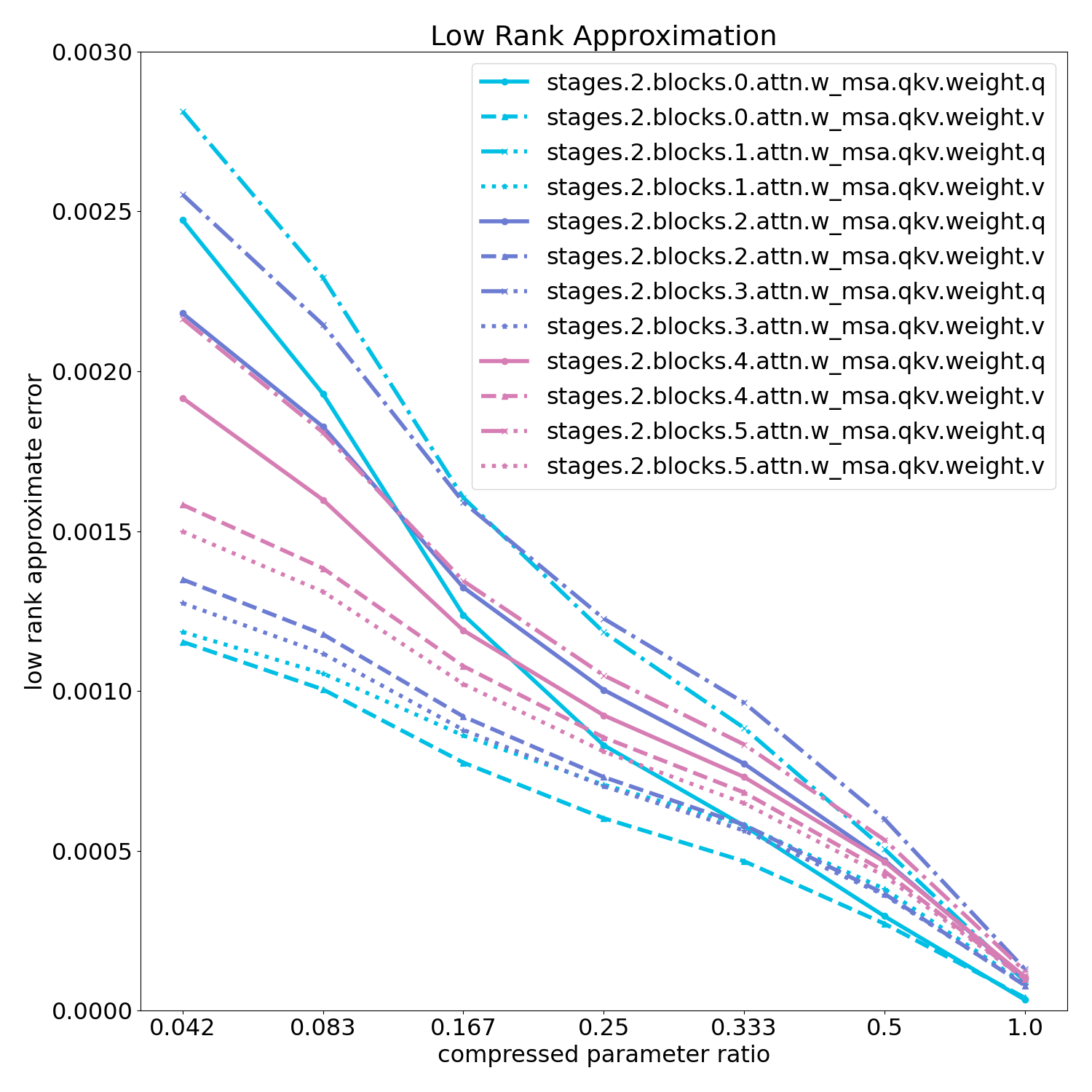}
   % \caption{backbone stage3}
   % \label{fig:sub3}
  \end{minipage}
  \hfill % 用于子图之间的间隙
  % 第二行第二个子图
  \begin{minipage}{0.485\linewidth}
    \centering
    \includegraphics[width=\linewidth]{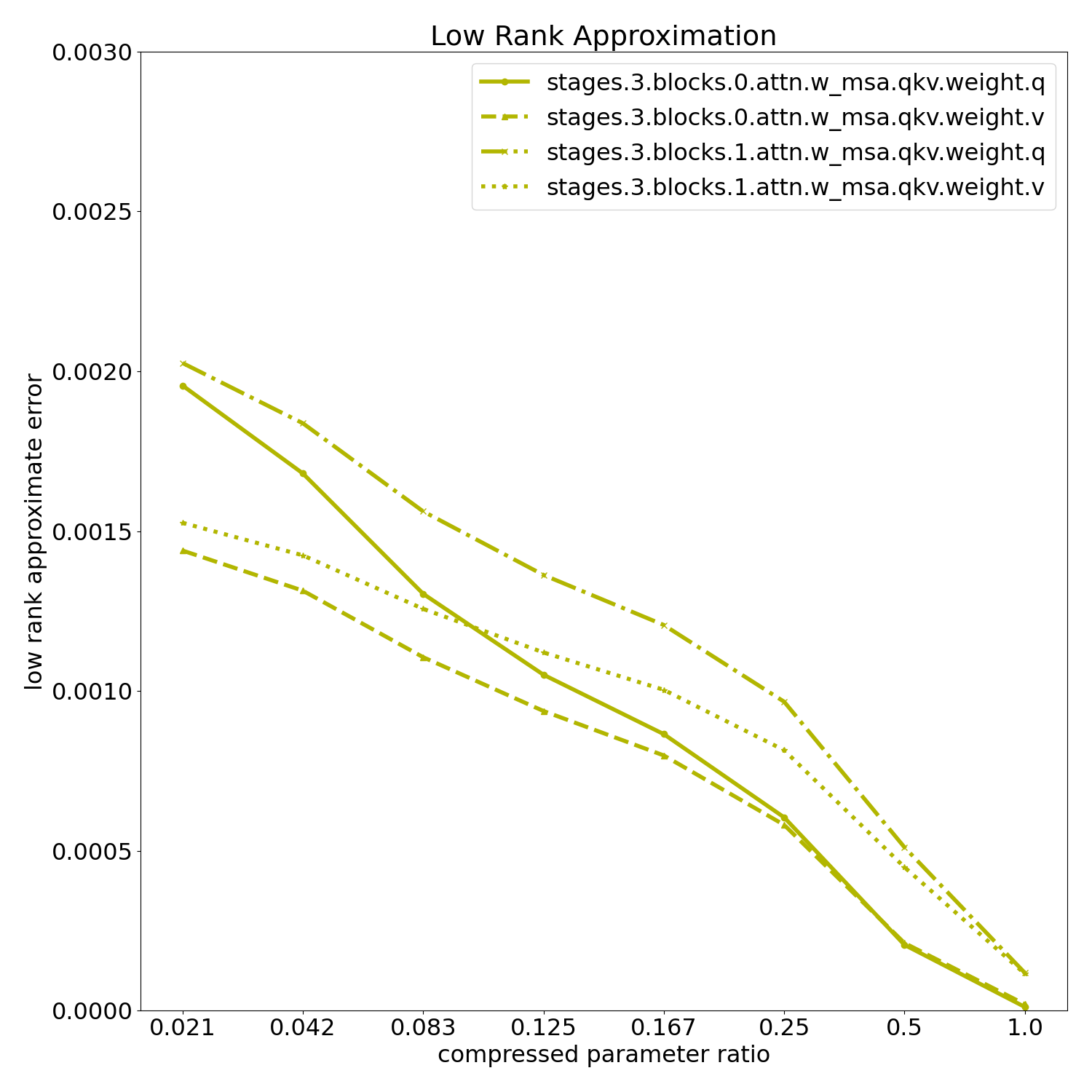}
   % \caption{backbone stage4}
   % \label{fig:sub4}
  \end{minipage}
  
  \caption{Low rank approximation of weight matrices on every stages on the Swin Transformer backbone, stage 1,2,3 and 4 for left-top, right-top, left-bottom and right-bottom sub-figures.}
  \label{fig:main}
\end{figure}

We further analyzed the relationship between the low-rank approximation error of the trainable parameter matrices in each stage of the Swin Transformer backbone and the compressed parameter ratio $p = (d \times r + r \times k)/(d \times k)$, in order to select an appropriate rank value $r$ for each module of LoRA-Det. As illustrated in Figure 10, four subfigures separately present the low-rank estimation results for stages 1, 2, 3, and 4 of the backbone. The x-axis represents the compressed parameter ratio $p$, with an increase in $p$ indicating a larger rank $r$ dimension of the low-rank decomposition, thus lower compression efficiency. The y-axis represents the low-rank approximation error. Experimental analysis reveals that when $p$ is small, that is, when a lower approximate rank dimension is adopted, the low-rank approximation error is significant; as $p$ increases, the error gradually decreases, only reaching a lower approximation error close to $p=1$. This indicates that the parameter matrices in the pre-trained weight of the backbone typically do not possess extremely low intrinsic dimensions and cannot be accurately approximated with a lower rank. Therefore, a larger low-rank value $r$ needs to be adopted in the LoRA module of the LoRA-Det backbone to ensure fine-tuning detection performance.

These experimental results are in line with the theoretical proofs in \cite{SAID}, which state that larger model weight matrices in deep learning networks tend to have low intrinsic dimensions. Analyzing the intrinsic dimensions of pre-trained weight parameter matrices through low-rank approximation methods helps us set the low-rank value $r$ in the LoRA module within LoRA-Det, design the distribution of trainable parameter quantities reasonably across different parts in the detection model, and achieve an ideal trade-off between model performance and parameter efficiency through minimal use of trainable parameters.

On the other hand, employing parameter efficient fine-tuning technologies on normal-size algorithms for downstream tasks is more challenging compared with large model due to the indeterminate intrinsic dimensions of weight matrices. Nearly full rank distributions of parameter matrices on relatively small-size model may lead to unsatisfactory PEFT results which are highly influenced by specific model frameworks. Therefore, carefully designed fine-tuning strategies and network architectures are crucial for the parameter efficient fine-tuning of general models and need urgent exploration in the future work.

\section{Conclusion}
To address the need for updating limited quantities of parameters in the satellite onboard models for improving detection performance, this paper presents a novel oriented object detection algorithm, LoRA-Det, which is based on hybrid parameter-efficient fine-tuning technologies including LoRA with low-rank approximation and full fine-tuning modules. Employing the elaborate PEFT mechanisms of multiple dense layers in the whole model architectures, it achieves highly competitive detection performance under the constraint of limited model parameter updates. The comprehensive experiments are conducted on three remote sensing image datasets to verify the efficiency of the proposed method. With only 12.4$\%$ of the model parameters being updated, it is capable of achieving 98$\%$ of the performance attained through full fine-tuning. Moreover, when compared with lightweight object detection networks having a similar volume of trainable parameters, the proposed method is validated to find an optimal balance between enhancing model performance and controlling the volume of parameters updated. The proposed LoRA-Det is able to leverage continuous data to improve the model's detection capabilities and generalization.

\bibliographystyle{IEEEtran}
	\bibliography{reference}

\vfill

\end{document}